%% file: main.tex
\pdfoutput=1
\documentclass[10pt,twocolumn,letterpaper]{article}

\usepackage[pagenumbers]{cvpr}
\usepackage{graphicx}

\input{preamble}

\definecolor{cvprblue}{rgb}{0.21,0.49,0.74}
\usepackage[pagebackref,breaklinks,colorlinks,allcolors=cvprblue]{hyperref}

\def\confName{CVPR}
\def\confYear{2027}

\title{Any2Act: Scaling In-Context Robot Learning\\from Heterogeneous Demonstrations}

\hypersetup{
  pdftitle={Any2Act: Scaling In-Context Robot Learning from Heterogeneous Demonstrations},
  pdfauthor={Ziang Ren, Zifan Wang, Zhuoyang Li, Benhao Qin, Haoyang Li, Qingtao Liu, Xuchuan Chen, Zekun Qi, He Wang, Yanchao Yang, Li Yi}
}

\newcommand{\aff}[1]{\textsuperscript{#1}}
\newcommand{\authsep}{\hspace{1.1em}}
\author{%
  \renewcommand{\arraystretch}{1.15}%
  Ziang Ren\aff{1,5,*}\authsep
  Zifan Wang\aff{2,5,*}\authsep
  Zhuoyang Li\aff{5,*}\authsep
  Benhao Qin\aff{5}\authsep
  Haoyang Li\aff{5}\authsep
  Qingtao Liu\aff{5}\\
  Xuchuan Chen\aff{2,5}\authsep
  Zekun Qi\aff{2,5}\authsep
  He Wang\aff{3,5}\authsep
  Yanchao Yang\aff{1,\textdagger}\authsep
  Li Yi\aff{2,4,5,\textdagger}\\[0.9em]
  {\normalsize
    \aff{1}The University of Hong Kong\authsep
    \aff{2}Tsinghua University\authsep
    \aff{3}Peking University\authsep
    \aff{4}Shanghai Qi Zhi Institute\authsep
    \aff{5}Galbot}\\[0.35em]
  {\small
    \aff{*}Equal contribution\authsep
    \aff{\textdagger}Corresponding authors}\\[0.75em]
  {\normalsize\href{https://any2act.github.io}{https://any2act.github.io}}
}

\begin{document}

\addtocontents{toc}{\protect\setcounter{tocdepth}{-10}}

\twocolumn[{%
    \renewcommand\twocolumn[1][]{#1}%
    \maketitle
    \centering
    \includegraphics[width=\textwidth, trim=0 120pt 0 20pt, clip]{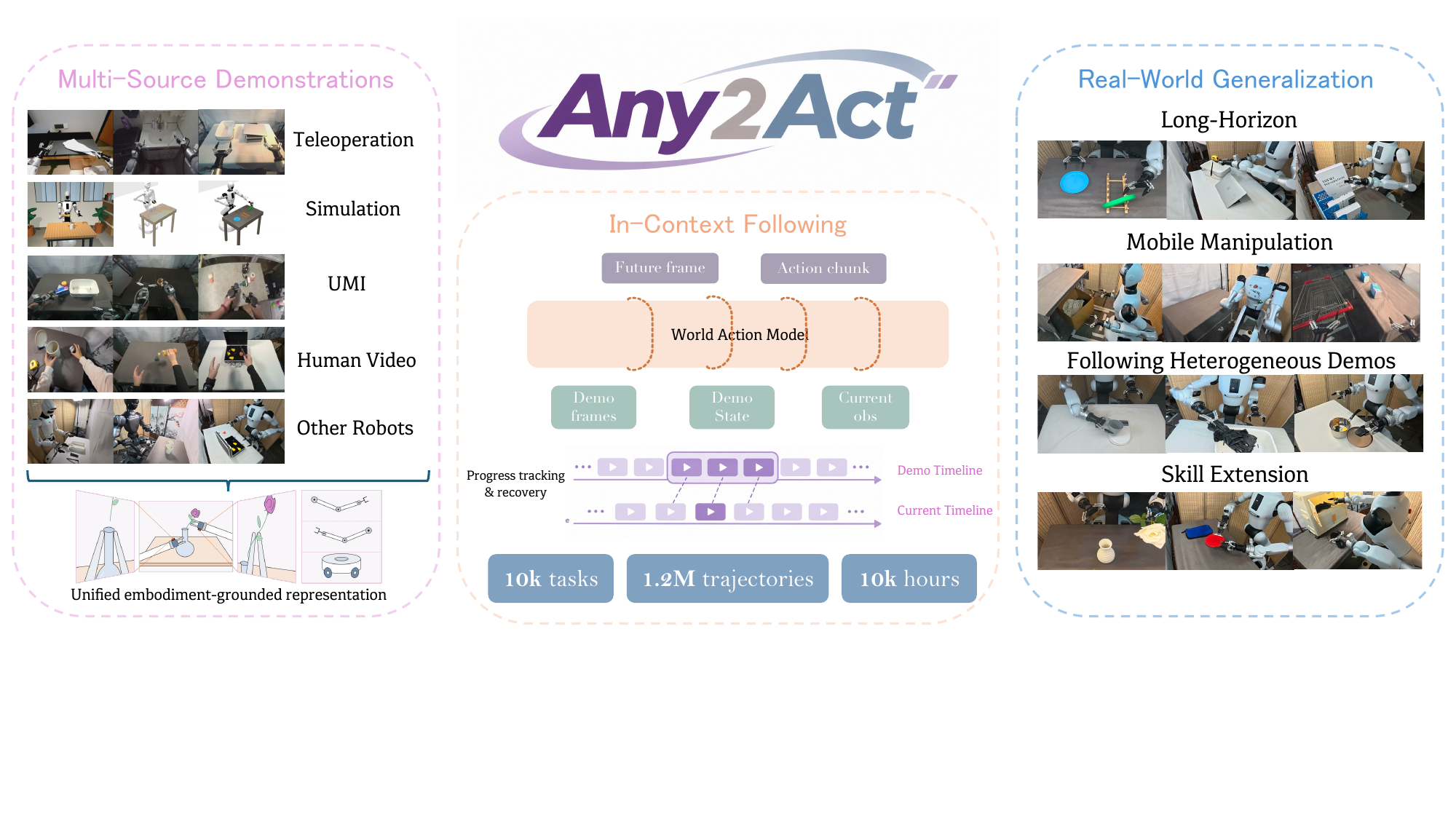}
    \captionsetup{hypcap=false}
    \captionof{figure}{%
       \textbf{Any2Act} lets an \emph{executor} robot follow any
       \emph{demonstrator}, whether teleoperator, simulator, UMI,
       human, or another robot, via a unified embodiment-grounded
       representation for real-world long-horizon mobile manipulation.%
    }%
    \label{fig:teaser}%
    \captionsetup{hypcap=true}%
    \vspace{4.5mm}
}]

\input{sec/0_abstract}
\input{sec/1_intro}
\input{sec/2_related_works}
\input{sec/3_method}
\input{sec/4_experiments}
\input{sec/5_conclusion}

{
    \small
    \bibliographystyle{ieeenat_fullname}
    \bibliography{main}
}

\clearpage
\appendixsetup
\addtocontents{toc}{\protect\setcounter{tocdepth}{1}}
\input{sec/A_appendix}

\end{document}

%% file: preamble.tex
\usepackage{multirow}
\usepackage{makecell}
\usepackage{tabularx}
\usepackage{algorithm}
\usepackage[noend]{algpseudocode}
\usepackage{xspace}
\usepackage[final]{microtype}

\newcommand{\ours}{Any2Act\xspace}
\newcommand{\taudem}{\hat{\tau}}
\newcommand{\para}[1]{\vspace{2pt}\noindent\textbf{#1}\hspace{0.4em}}

\newcommand{\appendixsetup}{%
  \appendix
  \setcounter{figure}{0}\setcounter{table}{0}%
  \setcounter{equation}{0}\setcounter{algorithm}{0}%
  \renewcommand{\thefigure}{S\arabic{figure}}%
  \renewcommand{\thetable}{S\arabic{table}}%
  \renewcommand{\theequation}{S\arabic{equation}}%
  \renewcommand{\thealgorithm}{S\arabic{algorithm}}%
  \def\theHfigure{S\arabic{figure}}%
  \def\theHtable{S\arabic{table}}%
  \def\theHequation{S\arabic{equation}}%
  \def\theHalgorithm{S\arabic{algorithm}}%
}
\AtEndPreamble{\AtBeginDocument{%
  \crefname{appendix}{App.}{Apps.}%
  \Crefname{appendix}{Appendix}{Appendices}%
  \crefname{subappendix}{App.}{Apps.}%
  \Crefname{subappendix}{Appendix}{Appendices}%
  \crefname{algorithm}{Alg.}{Algs.}%
  \Crefname{algorithm}{Algorithm}{Algorithms}%
}}

\makeatletter
\def\@maketitle{%
  \newpage
  \null
  \vskip .2in
  \begin{center}
    {\Large\bf \@title\par}
    \vspace*{18pt}
    {\large\lineskip .5em
      \begin{tabular}[t]{c}\@author\end{tabular}\par}
    \vspace*{10pt}
  \end{center}}
\makeatother

\makeatletter
\renewcommand*\l@section[2]{\ifnum \c@tocdepth >\z@ \addpenalty\@secpenalty \addvspace{0.5em \@plus\p@}%
  \setlength\@tempdima{1.6em}\begingroup \parindent \z@ \rightskip \@pnumwidth \parfillskip -\@pnumwidth
  \leavevmode \bfseries \advance\leftskip\@tempdima \hskip -\leftskip #1\nobreak\hfil \nobreak\hb@xt@\@pnumwidth{\hss #2}\par \endgroup\fi}
\makeatother

%% file: sec/0_abstract.tex
\begin{abstract}
Robot in-context learning (ICL) enables a policy to perform unseen tasks by following contextual demonstrations, without task-specific fine-tuning. This is increasingly valuable as large language models reasoning at low frequency can specify tasks through demonstrations and delegate high-frequency execution to such a policy. However, methods conditioned on human videos are bounded by scarce human--robot pairs, while those conditioned on robot demonstrations accept only one or two source types at limited scale. We present \textbf{Any2Act}, a foundation model for robot ICL. Our key insight is that a demonstration should be expressed in the camera layout and action space of the robot that performs it (the \emph{executor}), not of whoever provided it (the \emph{demonstrator}). This unified embodiment-grounded representation decouples learning to follow from admitting a new source. Any2Act learns to follow once, from robot-to-robot pairs that already take this form in 1.2M real trajectories spanning 10K tasks. Human videos, simulation, UMI, and other robots are then translated into this representation, and source-routed adapters trained on only a few hundred pairs each absorb the residual gap while the pretrained model stays frozen. For long horizons, a training-free tracker feeds only the demonstration window around the executor's progress, bounding the context and rolling back to redo failed steps. From a single demonstration, Any2Act achieves 57.5\% success on unseen tasks without fine-tuning, exceeding VLAs fine-tuned on 50 demonstrations per task; other demonstrators retain 74--91\% of the success achieved with the executor's own demonstrations.
\end{abstract}

%% file: sec/1_intro.tex
\section{Introduction}
\label{sec:intro}

Humans can acquire new skills simply by watching video demonstrations, a capability that robot learning has long aspired to. Toward this goal, robot in-context learning (ICL) has recently emerged~\cite{duan2017one,icrt,instantpolicy,jain2024vid2robot,host,zerowam}: conditioned on a few contextual examples, such as robot trajectories or human videos, a policy can directly perform unseen tasks that are semantically consistent with the given context, without any task-specific fine-tuning. This capability becomes even more important as large language models (LLMs) acquire stronger embodied reasoning abilities~\cite{saycan,codeaspolicies,team2026systematically}. Since such models typically reason and plan at a low frequency, a promising paradigm is to invoke an in-context policy as a \textit{tool call}: an LLM agent specifies the task through diverse demonstrations, and the policy executes it at a high control frequency. Serving in this role places stringent demands on the policy's generalizability and scalability.

However, existing ICL methods remain far from meeting these demands, facing three key challenges.
The first is \textbf{scalable supervision} for demonstration following. Methods conditioned on human videos must learn from task-aligned human–robot pairs~\cite{jain2024vid2robot,host,zerowam}, which remain scarce, and the videos themselves lack action labels and often hide hand–object contact behind the hand. Methods conditioned on robot demonstrations avoid such pairing, but are trained on synthetic or benchmark-scale data~\cite{instantpolicy,qian2026synthicl,patel2026bpp,ding2026contextflow}, which limits the range of tasks they can follow.
The second is \textbf{demonstration heterogeneity}. Each prior method accepts only one or two types of demonstration. In real-world deployment, however, a policy must handle not only diverse scenes and tasks but also heterogeneous demonstrations, such as human videos, simulation rollouts, UMI recordings~\cite{umi}, and trajectories from robots of different embodiments, and admitting a new type should not require a new large-scale paired dataset and retraining.
The third is \textbf{generalizable demonstration following}. A demonstration specifies not only what to do but also how, such as the grasp, the order of steps, and the bimanual coordination. Following it thus requires more than replay: the policy must transfer this motion to new objects and scenes, and, when execution deviates or fails, retry and resume over long horizons.

In this work, we present \textbf{Any2Act} (Fig.~\ref{fig:teaser}), a foundation model for robot in-context learning that addresses these challenges. We call the robot performing the task the \emph{executor}, and whatever provides the demonstration (a teleoperated robot, simulator, UMI, human, or another robot) the \emph{demonstrator}. Our key insight is to express a demonstration in the executor's camera layout and action space rather than the demonstrator's; we term this a \emph{unified embodiment-grounded representation}. A single policy can then follow demonstrations from any source, with lightweight adapters bridging the residual gap in appearance and motion. This decouples learning demonstration following, which needs large-scale paired data, from admitting a new source, which needs only a translation pipeline and a few hundred pairs.

To learn demonstration following at scale, we pretrain Any2Act on robot-to-robot demonstration pairs. Teleoperated collection records each task many times, and any two recordings of the same task form a pair with accurate actions and wrist and head views. Our corpus comprises 1.2M real trajectories (10K hours) spanning 10K tasks. No other demonstrator is involved at this stage: the following ability is learned once, in the executor’s own space, and is shared by every source later translated into it.

To handle demonstration heterogeneity, we admit new sources in two steps. First, every source is translated into the unified embodiment-grounded representation. For example, an agentic \textit{ego2sim} pipeline re-enacts a human video on a simulated executor, completing the wrist views and recovering the actions that egocentric videos lack. Second, to absorb the residual gap in appearance and motion without compromising the foundation model, we attach a plug-in source-routed adapter to each source, activated only on its translated yet still imperfect demonstration tokens. We show that a few hundred paired demonstrations suffice to adapt to a new source, while the base model's behavior remains unchanged. Paired data are thus needed only at this small scale per source, rather than to learn following itself.

For generalizable demonstration following, Any2Act reads not the whole demonstration as global context but a window around the executor's current progress, located by a training-free tracker, keeping the context focused and bounded over long horizons. This window also enables recovery: a failure such as a missed grasp or a slipped object often returns the scene close to an earlier phase, so the tracker rolls the window back to redo the failed step.

In real-world experiments, Any2Act performs unseen tasks with 57.5\% success without fine-tuning, exceeding VLAs fine-tuned on 50 demonstrations per task, while prior in-context policies reach at most 19.7\%. Demonstrations from simulation, human videos, UMI, and other robots retain 74--91\% of the success achieved with the executor's own demonstrations, each through adapters trained on only 200 pairs. We contribute: (1) Any2Act, a foundation model for demonstration following, pretrained at scale; (2) a unified embodiment-grounded representation with source-routed adapters, which admits heterogeneous demonstrations without changing the pretrained model; and (3) a training-free progress tracker that bounds the context over long horizons and enables failure recovery.

%% file: sec/2_related_works.tex
\section{Related Work}
\label{sec:relatedworks}

\begin{table*}[t]
\centering
\scriptsize
\setlength{\tabcolsep}{2.5pt}
\renewcommand{\arraystretch}{1.15}
\caption{\textbf{Comparison with open-source robot in-context learning policies.}
We report the data used for pretraining and mid-training, excluding additional data used to fine-tune each model for a specific evaluation domain. $^{\mathrm{syn}}$: synthetic data; /: unused or unreported.}
\label{tab:icl_comparison}
\begin{tabular*}{\textwidth}{@{\extracolsep{\fill}}lll ccccc c l@{}}
\toprule
\multirow{2}{*}{Method}
& \multicolumn{2}{c}{Pretrain / Mid-train Data}
& \multicolumn{5}{c}{Demo Sources for Real-Robot Inference}
& \multicolumn{2}{c}{Demo Conditioning} \\
\cmidrule(lr){2-3}\cmidrule(lr){4-8}\cmidrule(lr){9-10}
& \multicolumn{1}{c}{Human--Robot Pairs}
& \multicolumn{1}{c}{Robot Trajectories}
& \makecell{Human\\Video} & UMI & Simulation
& \makecell{Robot\\Same Emb.} & \makecell{Robot\\Cross Emb.}
& \makecell{w/\\State/Action} & Injection Mechanism \\
\midrule
SynthICL~\cite{qian2026synthicl}
& / & 3M$^{\mathrm{syn}}$
& & & & $\checkmark$ & &
& Global frames + cross-attn. \\
BPP~\cite{patel2026bpp}
& / & /
& & $\checkmark$ & & $\checkmark$ & & $\checkmark$
& Global memory + cross-attn. \\
HOST~\cite{host}
& 5,847 & 193,462 (229 tasks)
& $\checkmark$ & & & $\checkmark$ & &
& Progress-guided window \\
Zero-WAM~\cite{zerowam}
& 71,435$^{\mathrm{syn}}$ (8.6K tasks) & $\approx$400K ($>$6K tasks)
& $\checkmark$ & & & & &
& Full-demo prefix \\
ContextFlow~\cite{ding2026contextflow}
& / & /
& & & & $\checkmark$ & & $\checkmark$
& Global context + blockwise attn. \\
SimpleICL~\cite{li2026simpleicl}
& / & /
& $\checkmark$ & & & & &
& Global context + cross-attn. \\
\midrule
\textbf{Any2Act}
& / & \textbf{1.2M (10K tasks)}
& $\checkmark$ & $\checkmark$ & $\checkmark$ & $\checkmark$ & $\checkmark$ & $\checkmark$
& \textbf{Tracked window + rollback} \\
\bottomrule
\end{tabular*}
\end{table*}

\subsection{World Action Models}
World action models (WAMs) jointly predict future states and robot actions, and have emerged as a new architecture for generalist robot policies~\cite{lu2026wamsurvey,wang2026world,li2026egowam,wang2026openwam}. With large-scale pretraining, they exhibit strong generalization: DreamZero~\cite{ye2026worldactionmodelszeroshot} generalizes zero-shot to unseen tasks after learning from non-repetitive robot data, Motus~\cite{bi2026motus} learns from human and multi-robot data via latents, and GR00T~2~\cite{nvidia2026gr00t2} transfers tasks zero-shot across embodiments. The architecture is also flexible, supporting unified policy and dynamics modeling~\cite{zhu2025uwm}, value-based planning~\cite{cosmos_cookbook_cosmos_policy_2026}, autoregressive closed-loop control~\cite{li2026lingbotva}, inference without future generation~\cite{yuan2026fastwam}, and steering by human videos~\cite{feng2026wam,zerowam}. Any2Act is likewise pretrained at scale, but in context: it learns to perform tasks specified by a demonstration. Whereas such pretraining usually absorbs heterogeneous data into model weights, Any2Act admits heterogeneous sources at inference time as demonstrations. To our knowledge, it is the first WAM to do so for multiple sources.

\subsection{In-Context Learning for Robot Manipulation}
Demonstrations can specify spatial and temporal details that language leaves ambiguous. Robot in-context learning uses them to guide execution without task-specific parameter updates~\cite{duan2017one,jain2024vid2robot,huang2026context,mosaic}. Recent industrial systems such as GEN-1.5~\cite{generalist2026gen15} and Skild S1~\cite{skild2026s1} perform new tasks from a single demonstration, highlighting the potential of this paradigm. Tab.~\ref{tab:icl_comparison} compares recent open-source policies along three axes. \emph{Training data:} prior methods rely on synthetic pairs~\cite{qian2026synthicl}, benchmark imitation data~\cite{patel2026bpp,ding2026contextflow}, human--robot pairs~\cite{host}, or generated human videos~\cite{zerowam}; Any2Act is pretrained on robot-to-robot pairs drawn from 1.2M real trajectories, which are collected naturally from routine data collection with accurate actions. \emph{Demonstration sources:} each prior method accepts one or two; Any2Act is the only one that admits all five, including simulation and cross-embodiment robots, through a unified embodiment-grounded representation. \emph{Conditioning:} prior methods encode the whole demonstration as global context~\cite{qian2026synthicl,patel2026bpp,ding2026contextflow,li2026simpleicl} or a full prefix~\cite{zerowam}, or track progress with a learned aligner, without demonstration states and rollback~\cite{host}; Any2Act pairs a training-free tracker that rolls back after failures with dense demonstration states, keeping the context bounded while conveying the motion.

%% file: sec/3_method.tex
\section{Method}
\label{sec:method}

\begin{figure*}[t]
    \centering
    \includegraphics[width=\textwidth]{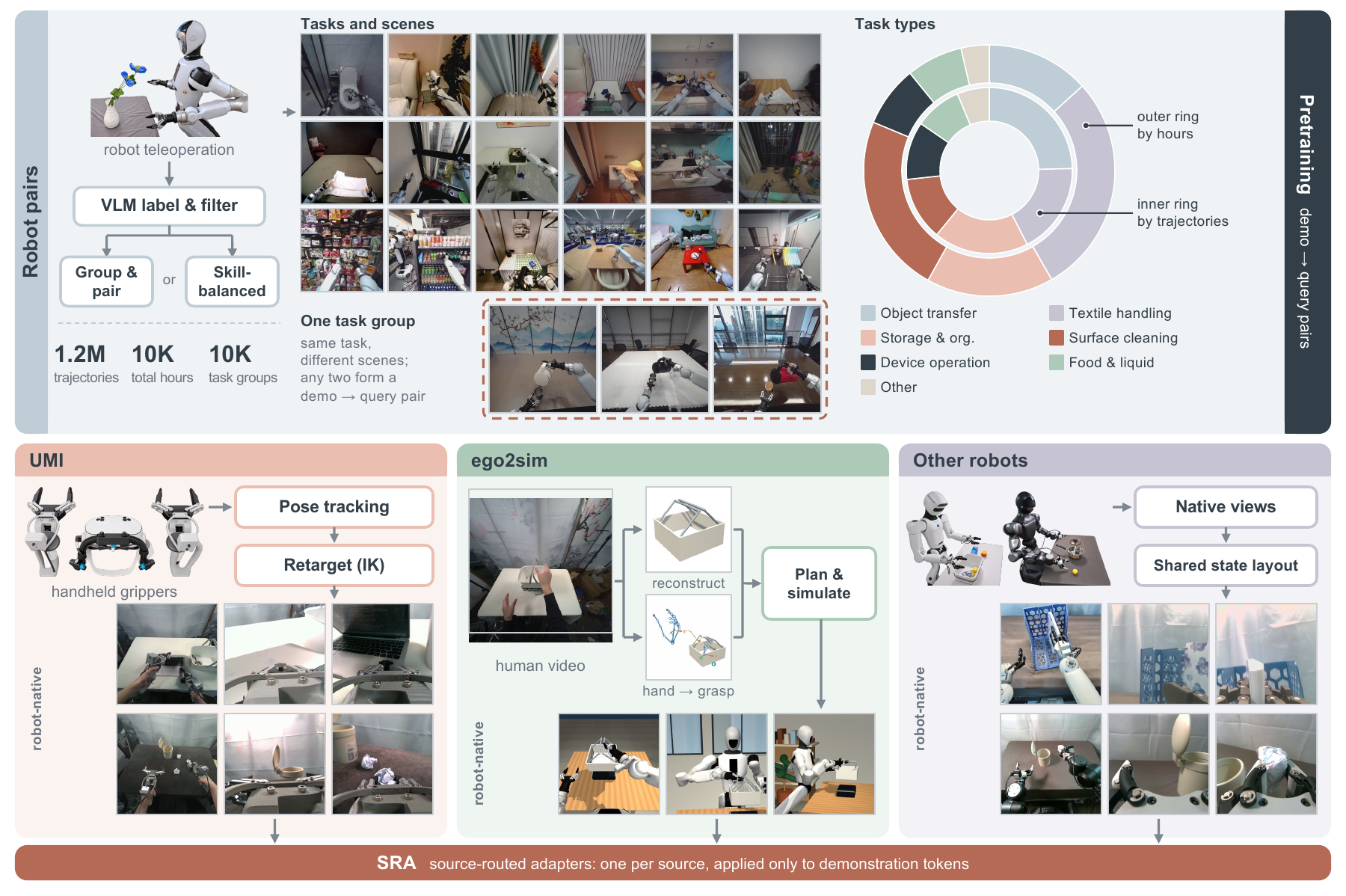}
    \caption{\textbf{Demonstration sources.} Top: robot-to-robot pairs from routine data collection
supply pretraining; recordings of the same task are VLM-filtered and grouped, and any two form a
pair. Bottom: human videos (via agentic ego2sim), UMI, and other robots are translated
into the unified embodiment-grounded representation and read through source-routed adapters.}
    \label{fig:data}
\end{figure*}

Any2Act drives a target robot, the \emph{executor}, to perform a task specified by a demonstration.
The \emph{demonstrator} that performed it, whether a person using bare hands or a handheld UMI
gripper, a simulated robot, or another real-world robot, need not share the executor's body.
Given the executor's current multi-view observation $o_t$, its proprioceptive history $q_{\le t}$,
a language instruction $\ell$, and a demonstration $\mathcal{D}$ of the intended task from source $k$,
the Any2Act policy $\pi$ generates, without task-specific parameter updates, an
$H$-step action chunk for the arms, grippers, and mobile base, together with a prediction of the
scene $h$ steps ahead:
\begin{equation}
\big(a_{t:t+H},\; o_{t+h}\big) \sim
\pi_{\theta,\phi_k}\big(\,\cdot \mid o_t,\, q_{\le t},\, \ell,\,
\mathcal{W}(\mathcal{T}_k(\mathcal{D}), o_{\le t})\,\big).
\label{eq:policy}
\end{equation}

Any2Act follows heterogeneous demonstrations in three steps. The translation $\mathcal{T}_k$ maps the demonstration into the executor's camera layout and action space, a unified embodiment-grounded representation (Sec.~\ref{sec:representation}). Progress tracking $\mathcal{W}$ then crops the window that matches the executor's current progress, and rolls it back when a step fails (Sec.~\ref{sec:align}). The model reads this window with the current observation and predicts the actions and the future frame (Sec.~\ref{sec:model}). Training proceeds in two stages: $\theta$ is pretrained once on robot-to-robot pairs, for which $\mathcal{T}_k$ is the identity and $\phi_k$ is empty (Sec.~\ref{sec:pretrain}), and each new source then adds only source-routed adapters $\phi_k$, with $\theta$ frozen (Sec.~\ref{sec:plugins}).

\subsection{Embodiment-Grounded Representation}
\label{sec:representation}

Demonstrations from different demonstrators differ in what they show and what they record.
They are captured from different views and bodies, such as human hands, handheld grippers,
rendered scenes, or other robots. Most also lack the executor's actions: a human video records none,
a UMI recording only end-effector poses, and another robot its own joint layout. Feeding them to the
policy as they are would require learning a separate mapping for each source from scarce paired data. To bridge these differences, we map every demonstration with a source-specific
translation $\mathcal{T}_k$ into a unified embodiment-grounded representation: a head view and two
wrist views that follow the executor's camera layout, paired with motion in the executor's action
space of the arms and  mobile base. Following is thus learned once in this space
(Sec.~\ref{sec:pretrain}) and shared by every source.

Apart from teleoperated recordings on the executor, which need no translation, demonstrations are
translated in three ways (Fig.~\ref{fig:data}). (1)~\emph{Simulation.} Following recent real-to-sim pipelines~\cite{zhao2025robot,li2026arstag}, we quickly model the current scene in a physics simulator from a head-camera image or video and an asset library; an LLM agent~\cite{team2026systematically} then rolls out the task on the simulated executor, generating its actions directly (\cref{app:sources_sim}). This yields views and actions in the executor's layout, and failed attempts can be retried freely, returning a verified demonstration in minutes; the cost is an appearance gap and inaccurate physics. (2)~\emph{Human videos.} Our agentic ego2sim pipeline (\cref{app:sources_human}) builds
on recent work that reconstructs egocentric hand--object interaction in
simulation~\cite{liu2026egoengine}: it estimates hand and object poses from the video,
rebuilds the scene, transfers the grasp to the robot gripper, and executes the human trajectory on
the simulated executor. This recovers the actions and wrist views that egocentric videos lack, at
the cost of errors in hand--object tracking and retargeting.
(3)~\emph{Other real-world bodies.} UMI recordings and other robots' data already contain a head
view and two wrist views, so only their motion needs to be mapped into the executor's action space,
from end-effector poses for UMI and joint layouts for other robots.

Translated demonstrations share the executor's format but not its distribution: UMI frames still show a handheld gripper, simulated rollouts carry rendering artifacts and inaccurate physics, and converted motion inherits tracking and kinematic errors. We absorb these residual differences with source-routed adapters (Sec.~\ref{sec:plugins}).

\subsection{Progress Tracking and Recovery}
\label{sec:align}

Once translated, a demonstration is followed step by step, so encoding it in full would make the context grow with its length. The policy instead reads only a short window around the executor's current progress. Placing this window requires knowing which demonstration step the executor has reached, which is hard in the real world: at deployment only the past is observed; the executor moves at a different speed, pauses, and deviates; and demonstrations come from other bodies and unseen tasks. Rather than a learned aligner~\cite{host}, which would need retraining for every source, we split alignment in two: a rough, training-free tracker only places the current moment inside the window, and the policy, trained to tolerate its errors, pinpoints its progress within it.

Concretely, at each execution step $t$ the tracker finds an anchor, the demonstration step that best
matches the executor's current observation, using frozen DINOv3~\cite{simeoni2025dinov3} features $f$.
Besides the head view, we match the wrist views, which add the gripper--object relation that the head
view often misses~\cite{ren2026fisheyevla}, especially across sources (Sec.~\ref{sec:exp_align}). The matching cost between
execution step $t$ and demonstration step $s$ is
$c_t(s)=\sum_{v} w_v\big(1-\cos(f(o^{v}_{t}),f(d^{v}_{s}))\big)$, where $o^{v}_{t}$ and $d^{v}_{s}$
are the executor's and the translated demonstration's frames from view $v$, weighted by $w_v$ (values in \cref{app:tracking_forward}).
Since progress normally moves forward, the anchor follows a monotonic path, as in dynamic time
warping~\cite{sakoe1978dtw}: between consecutive execution steps it may stay, while the executor pauses, or advance by at
most $K$ demonstration steps; a penalty $\lambda(\delta)$ on the advance $\delta$ discourages stalling
and skipping. Let $F_t(s)$ be the lowest accumulated cost of such a path over execution steps
up to $t$ and ending at demonstration step $s$, and $B_t(s)$ the lowest cost of continuing it to the end
of the execution:
\begin{equation}
\begin{aligned}
F_t(s) &= c_t(s) + \min_{0\le\delta\le K}\big[\,F_{t-1}(s-\delta)+\lambda(\delta)\,\big],\\
B_t(s) &= \min_{0\le\delta\le K}\big[\,c_{t+1}(s+\delta)+\lambda(\delta)+B_{t+1}(s+\delta)\,\big].
\end{aligned}
\label{eq:align}
\end{equation}
During training, the full execution is in view, and the anchor $\tau_t=\arg\min_s[F_t(s)+B_t(s)]$
lies on the globally optimal path; it is computed once per pair, offline (\cref{app:tracking_tolerance}). At deployment only the past is observed, so the tracker gives an
approximate anchor $\hat\tau_t=\arg\min_s F_t(s)$, and $\mathcal{W}$ returns the demonstration window
$[\hat\tau_t-W_{\mathrm{p}},\,\hat\tau_t+W_{\mathrm{f}}]$. To cover the deviation of $\hat\tau_t$ from $\tau_t$ and differences in speed, during training the
window is centered at a perturbed anchor $\tau_t+\epsilon$, with $\epsilon$ spanning the typical
deviation while keeping $\tau_t$ inside the window, and randomly stretched or compressed in time;
the policy thus learns to locate its own progress by comparing the demonstration with the current
observation.

Because the policy reads only the window, the anchor acts as its pointer into the demonstration,
which also enables recovery. The policy itself absorbs small deviations, such as a displaced object, since pretraining pairs already differ in layout. A failure, however, often
returns the scene to a state the demonstration has already shown: after a missed grasp, the object
stays around the original place; after a slip, it lies back on the table. Since the
monotonic tracker does not move back by itself, we re-localize the anchor to an earlier phase when
it explains the current observation clearly better (details in \cref{app:rollback}), so the
robot redoes the failed step without recovery data or additional training.

\begin{figure*}[t]
    \centering
    \includegraphics[width=\textwidth, trim=0 8pt 0 0.5cm, clip]{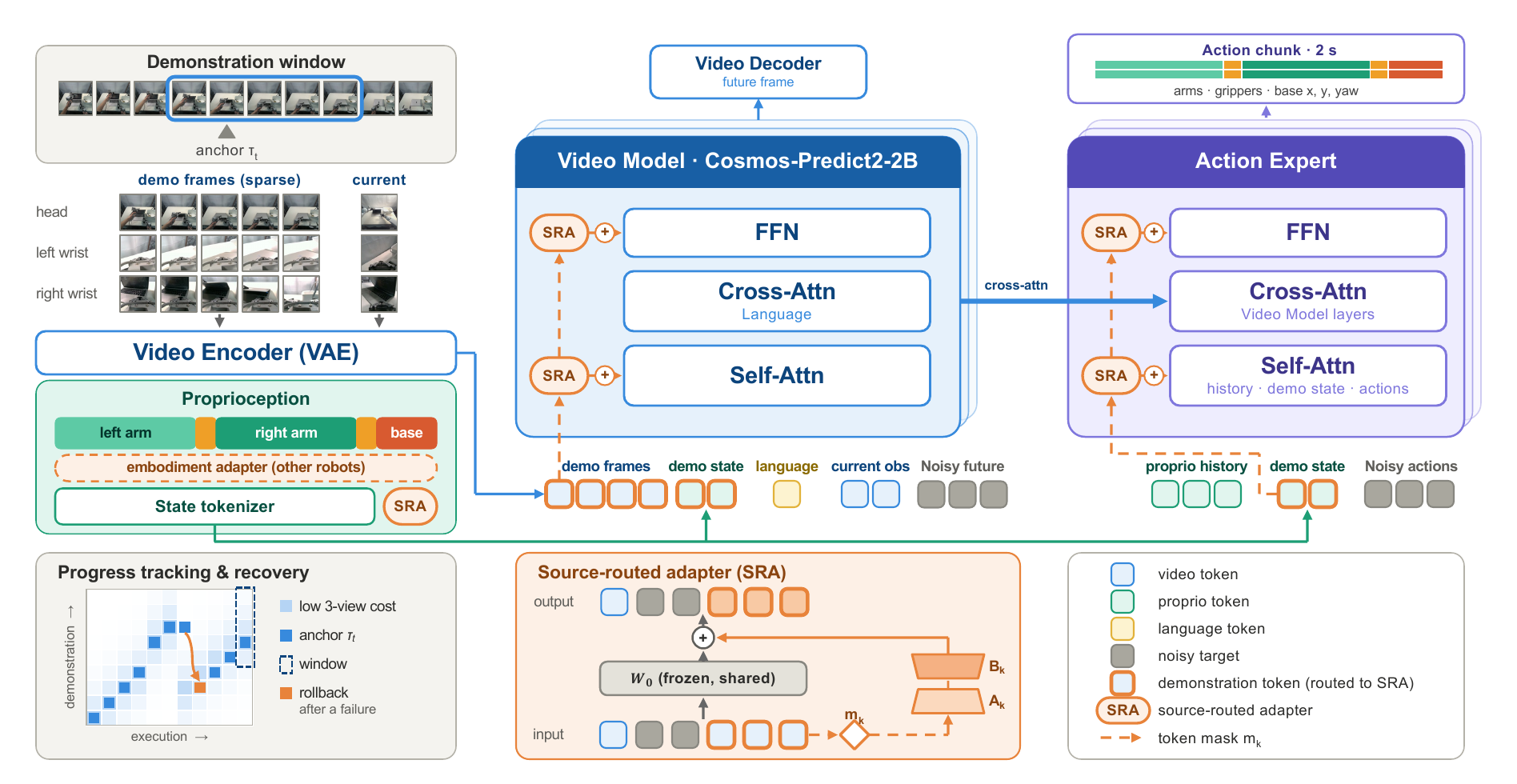}
    \caption{\textbf{Model overview.} Top: from a demonstration window, given as sparse multi-view
frames and dense proprioception, together with the current observation and the language, the video model predicts the future frame, and the action expert,
cross-attending to its layers, denoises the action chunk.
Bottom left:
progress tracking advances the window along the execution and rolls it back to an earlier phase
after a failure. Bottom middle: source-routed adapters (SRA) act only on demonstration tokens,
leaving the pretrained model unchanged.}
    \label{fig:model}
\end{figure*}

\subsection{Model Architecture}
\label{sec:model}

Given the demonstration window and the current observation, Any2Act predicts how the scene should evolve and the actions that make it so (Fig.~\ref{fig:model}). This follows world action models~\cite{ye2026worldactionmodelszeroshot,zhu2025uwm}, which predict future video alongside actions so that the priors of video pretraining carry over to action generation. In context, this prediction is harder and more important: the future must follow the demonstration while fitting a current scene with a different layout and objects. We therefore build Any2Act on the pretrained video model Cosmos-Predict2~\cite{cosmos} and cast demonstration following as conditional continuation.

The demonstration window is given in two complementary forms: sparse visual tokens, encoded from a
few frames of the three views to convey the scene, and dense state tokens, summarizing the
proprioception over the window to convey motion and timing. The executor is given likewise, by its
current views and its proprioceptive history; the language instruction enters through cross-attention.
\Cref{fig:tokens} details the token layout.

Two streams produce the two outputs. The video model places the demonstration tokens and the current
views in one sequence, so that the two can be compared directly, and predicts the future frame. A
lightweight action expert takes the executor's proprioceptive history and the demonstration state
tokens, reads the video model's features through cross-attention, and predicts the action chunk.
The two outputs are generated jointly to stay consistent. At deployment, demonstration latents are
precomputed and the context is encoded once per control step and reused across denoising steps,
yielding a two-second 30\,Hz action chunk in 180\,ms on one H200 GPU (\cref{app:model_infer}).

\subsection{Demonstration-Following Pretraining}
\label{sec:pretrain}

Learning to follow demonstrations at all requires numerous demonstration-following pairs. Human--robot pairs with accurate action labels are scarce, so prior work resorts to
synthetic human videos~\cite{zerowam} or a few thousand real pairs~\cite{host}. Our unified
representation removes the need for such pairs: the policy only needs to learn to follow
demonstrations within it, and robot-to-robot pairs already come in this form, with $\mathcal{T}_k$
the identity. Such pairs are also easy to obtain at scale: teleoperated data collection usually
records each task many times, and any two recordings of the same task form a pair with accurate
actions and all three views.

Our corpus contains 1.2M trajectories over 10K hours in 10K tasks
(Fig.~\ref{fig:data}), labeled and filtered by a VLM~\cite{qwen38} and grouped by task. Recordings of the same task
span different environments, object instances, and placements, so the policy must carry the
demonstrated behavior over to the present scene rather than replay it. It is the demonstration, not
a memorized task identity, that tells the policy what to do: removing it sharply degrades
performance (Sec.~\ref{sec:exp_ablation}). We pretrain $\theta$ end to end on the
perturbed windows of Sec.~\ref{sec:align}, with language randomly dropped so that the demonstration
remains the primary task specification. The pair construction and data sampling
(\cref{app:data_pairs}), the training objective (\cref{app:model_expert}), and all hyperparameters
(\cref{tab:hparams}) are detailed in the appendix.

\subsection{Source-Routed Adapters for New Sources}
\label{sec:plugins}

With $\theta$ pretrained on robot pairs, a demonstration translated from a new source already arrives in the executor's format, yet its residual differences in appearance and motion (Sec.~\ref{sec:representation}) change how the policy should interpret it. Fine-tuning $\theta$ directly is risky: paired data for a new source are scarce, and fully fine-tuning on so few pairs risks overfitting and degrading the demonstration-following ability learned from robot data (\cref{tab:sft_vs_sra}). We therefore keep $\theta$ frozen and propose \emph{source-routed adapters} (SRAs), which adapt only how the policy interprets each source's demonstrations, using a few hundred pairs per source.
Each SRA is a low-rank residual branch~\cite{lora} attached to the frozen model and activated only on the demonstration tokens of its source $k$:
\begin{equation}
h' = W_0 h + m_k\, B_k A_k h,
\label{eq:sra}
\end{equation}
where $h$ is a token feature, $W_0$ is a frozen linear layer, $A_k$ and $B_k$ are the low-rank
factors of source $k$ at that layer, which over all layers form $\phi_k$, and $m_k \in \{0,1\}$
indicates whether the token belongs to a demonstration of source $k$. Adapters are inserted into
layers of both the video model and the action expert, covering the visual and the state tokens of
the demonstration.
For other robots, whose kinematics also differ, an additional embodiment adapter is trained on the
same pairs (\cref{app:sra_emb}).
Demonstrations recorded on the executor itself pass through no adapter, so
following them behaves exactly as before adaptation. Adapters of different sources act on different
demonstrations and can thus be attached, removed, or combined without retraining. We show that each
new source needs only a few hundred pairs to gain the ability to
follow its demonstrations (Sec.~\ref{sec:exp_sra}).

%% file: sec/4_experiments.tex
\section{Experiments}
\label{sec:exp}

We first specify how Any2Act is compared with VLAs and prior in-context policies (Sec.~\ref{sec:compare_protocol}).
We then compare it with these methods in the real world (Sec.~\ref{sec:exp_real}) and on RoboTwin~2.0 (Sec.~\ref{sec:exp_robotwin}), and examine how its in-context ability scales with robot-to-robot pretraining and which components it relies on (Sec.~\ref{sec:exp_ablation}).
Finally, we turn to other demonstrators: we test whether simulation, human videos, UMI recordings, and other robots can specify tasks once translated and read through source-routed adapters (Sec.~\ref{sec:exp_sra}), and how accurately the training-free tracker locates progress (Sec.~\ref{sec:exp_align}).
Unless stated otherwise, all real-world experiments run on the same dual-arm mobile robot, in scenes unseen during pretraining, and each success rate is measured over 100 trials per task.

\begin{table*}[t]
\centering\footnotesize\setlength{\tabcolsep}{2.4pt}
\caption{\textbf{Real-world success rate (\%)}. In-context policies run without fine-tuning on Task Generalization and are adapted with 20 demonstrations per task on Skill Extension; VLAs are fine-tuned on 50 per task on both. $^\dagger$\,Long-horizon ($\geq$5 steps).}
\label{tab:main}
\begin{tabular}{l|cccccc|c|cccccc|c|c}
\toprule
 & \multicolumn{7}{c|}{Task Generalization} & \multicolumn{7}{c|}{Skill Extension} & \\
Method & {\scriptsize Discarding} & {\scriptsize Lidding} & {\scriptsize Gathering} & {\scriptsize Pinching} & {\scriptsize Filling$^\dagger$} & {\scriptsize Packing$^\dagger$} & Avg & {\scriptsize Relaying$^\dagger$} & {\scriptsize Sorting$^\dagger$} & {\scriptsize Storing$^\dagger$} & {\scriptsize Bagging} & {\scriptsize Extracting$^\dagger$} & {\scriptsize Pouring$^\dagger$} & Avg & All\\
\midrule
$\pi_{0.5}$~\cite{pi05} & 37 & 41 & 43 & 24 & \textbf{57} & 56 & 43.0 & \textbf{63} & 19 & 48 & 29 & 61 & 39 & 43.2 & 43.1\\
GR00T N1.7~\cite{gr00t_n17} & 62 & 33 & 57 & 28 & 44 & 61 & 47.5 & 42 & 39 & 67 & 45 & 69 & 23 & 47.5 & 47.5\\
HOST~\cite{host} & 19 & 26 & 0 & 13 & 11 & 25 & 15.7 & 13 & 33 & 35 & 13 & 31 & 25 & 25.0 & 20.3\\
Zero-WAM~\cite{zerowam} & 24 & 16 & 17 & 19 & 8 & 34 & 19.7 & 21 & 46 & 65 & 7 & 33 & 27 & 33.2 & 26.4\\
Any2Act & \textbf{64} & \textbf{47} & \textbf{77} & \textbf{38} & 56 & \textbf{63} & \textbf{57.5} & 61 & \textbf{88} & \textbf{79} & \textbf{55} & \textbf{84} & \textbf{69} & \textbf{72.7} & \textbf{65.1}\\
\bottomrule
\end{tabular}
\end{table*}

\begin{figure*}[t]
    \centering
    \includegraphics[width=\textwidth]{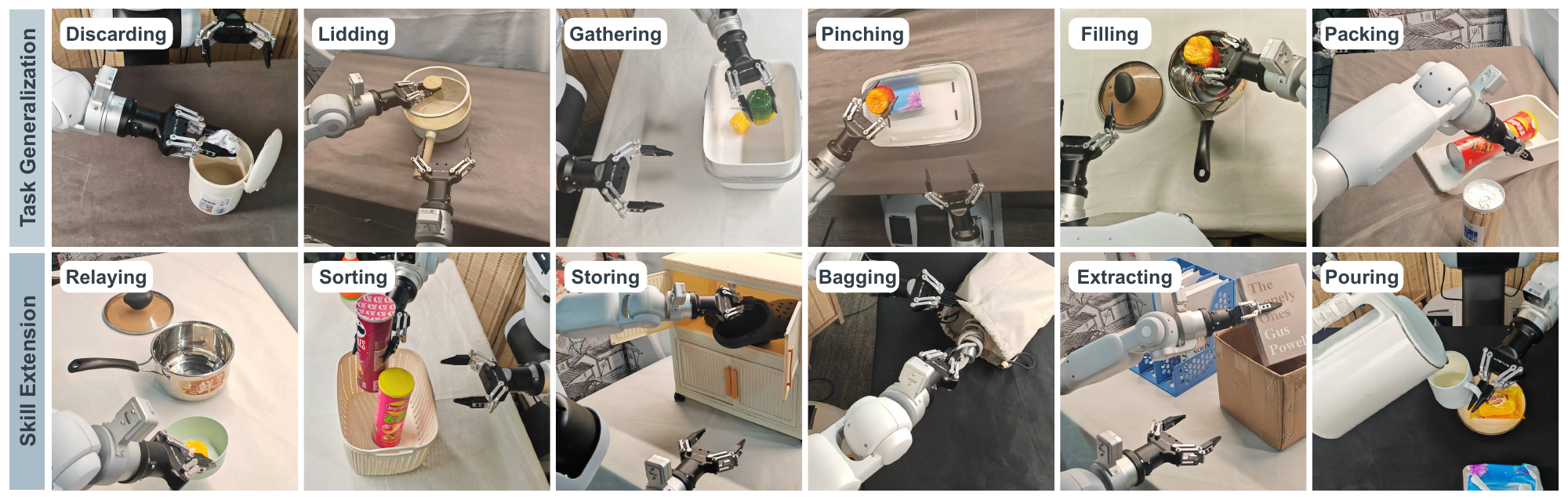}
    \caption{\textbf{Real-world benchmarks.} Task Generalization (top) composes pretrained skills into new tasks, while Skill Extension (bottom) requires skills beyond pretraining. None of the twelve tasks appears in pretraining, and all are evaluated in unseen scenes.}
    \label{fig:exp}
\end{figure*}

\subsection{Comparison Protocol}
\label{sec:compare_protocol}
Comparing in-context policies with prior methods is subtle, because they differ in what they receive at test time and in what they must be trained on beforehand.
Language-conditioned VLAs need no demonstration at inference, but acquire a new task only through fine-tuning on data from the same environment; we therefore fine-tune them on every evaluation task.
In-context policies need no task-specific fine-tuning in principle, but their generalization is bounded by the tasks, skills, visual inputs, and embodiment they were trained on, so their released checkpoints cannot be evaluated directly on our robot.
We therefore retrain them with their own methods on all of our robot pairs, and evaluate them with their required demonstrations and no fine-tuning.
Some settings, however, lie beyond what any pretraining corpus covers: a simulator differs from the real world in appearance and physics, and some tasks require skills or object properties absent from pretraining, such as deformable or articulated objects.
A single demonstration specifies what to do but cannot supply such a skill.
In these settings, all in-context policies, including ours, are first adapted by supervised fine-tuning on a small in-domain set, and are still conditioned on a demonstration at test time.

\subsection{Evaluating in the Real World}
\label{sec:exp_real}

\noindent\textbf{Platform and benchmark.}
All real-world experiments run on a Galbot G1~\cite{galbot_g1} dual-arm mobile robot with a head camera and two wrist cameras.
We construct two benchmarks of six tasks each (Fig.~\ref{fig:exp}; descriptions and success criteria in \cref{tab:tasks}), none of which appears in pretraining.
Both span a range of difficulty; they differ in whether the required skills are covered by pretraining.
The \emph{Task Generalization} benchmark tests whether a policy can perform a new task from a demonstration alone: its tasks compose skills largely seen in pretraining into new goals, step orders, and object combinations, and Any2Act executes them without fine-tuning.
The \emph{Skill Extension} benchmark tests whether a policy can acquire skills beyond pretraining, such as handling deformable objects or atomic actions never seen before; here every in-context policy is first adapted as in Sec.~\ref{sec:compare_protocol}.
In both benchmarks, object instances, layouts, and backgrounds differ substantially from those in pretraining, adaptation, and the given demonstration, so the policy must transfer the demonstrated behavior to a new scene.

\noindent\textbf{Baselines.}
We compare with two language-conditioned VLAs, $\pi_{0.5}$~\cite{pi05} and GR00T N1.7~\cite{gr00t,gr00t_n17}, and two in-context policies, HOST~\cite{host} and Zero-WAM~\cite{zerowam}.
The VLAs are fine-tuned on 50 demonstrations per task on both benchmarks, since 20 proved insufficient for them (\cref{app:exp_baselines}).
HOST and Zero-WAM are retrained on our robot pairs, and like ours, run without fine-tuning on Task Generalization and are adapted with 20 demonstrations per task on Skill Extension.
We report success rate (\%) over 100 trials per task (protocol in \cref{app:exp_benchmark}).

\noindent\textbf{Evaluation.}
Table~\ref{tab:main} reports the results; \cref{tab:main_pertask} adds confidence intervals.
On \textit{Task Generalization}, Any2Act reaches 57.5\% without any fine-tuning, above both VLAs even though they are fine-tuned on 50 demonstrations of each task.
We observe a clear gain on \emph{Gathering}, a multi-step task, and \emph{Pinching}, which requires an unusual grasp: the demonstration conveys the step sequence and the intended grasp, which a language instruction leaves implicit.
On \textit{Skill Extension}, Any2Act reaches 72.7\% after adaptation with 20 demonstrations per task, against 47.5\% for GR00T N1.7 fine-tuned on 50.
The gain is largest on the long-horizon \emph{Sorting} and the bimanual \emph{Pouring}, where the demonstration specifies the order of steps and the coordination between the two arms.
HOST and Zero-WAM are trained on the same robot pairs and given the same demonstrations, yet reach only 15.7\% and 19.7\% on \textit{Task Generalization}, so the gain comes from how Any2Act uses the demonstration rather than from the training data.

\subsection{Evaluating in Simulation}
\label{sec:exp_robotwin}

\begin{table}[t]
\centering\small\setlength{\tabcolsep}{3pt}
\caption{\textbf{Success rate (\%) on seven unseen RoboTwin~2.0 tasks.}}
\label{tab:robotwin}
\begin{tabular*}{\columnwidth}{@{\extracolsep{\fill}}@{}lccc@{}}
\toprule
Task & \makecell{WAN-Action\\\cite{zerowam}} & \makecell{Zero-WAM\\\cite{zerowam}} & Any2Act\\
\midrule
Place object scale  &  3.0 & 24.7 & \textbf{59.3}\\
Stamp seal          &  7.3 & 47.0 & \textbf{61.7}\\
Open microwave      &  2.3 & \textbf{59.0} & 25.3\\
Move stapler pad    & 10.7 & 69.1 & \textbf{80.3}\\
Place bread basket  & 15.3 & 35.0 & \textbf{37.7}\\
Place empty cup     & 38.3 & 84.9 & \textbf{87.7}\\
Stack blocks three  &  0.0 &  9.0 & \textbf{20.0}\\
\midrule
Average             & 11.0 & 47.0 & \textbf{53.1}\\
\bottomrule
\end{tabular*}
\end{table}

We follow the cross-task protocol of Zero-WAM~\cite{zerowam}
on RoboTwin~2.0~\cite{robotwin2}, with 43 seen tasks for
RoboTwin post-training and 7 unseen tasks for evaluation.
Any2Act receives a simulation video for each unseen
task.
Success rate is averaged over three seeds of 100 rollouts, and baseline numbers are taken from~\cite{zerowam}; \cref{app:exp_robotwin} details the protocol.
As shown in Table~\ref{tab:robotwin}, Any2Act reaches 53.1\% on average, against 47.0\% for Zero-WAM and 11.0\% for its language-conditioned counterpart WAN-Action.

\subsection{Scaling and Ablations}
\label{sec:exp_ablation}

We next examine where this in-context ability comes from.
We pretrain the same configuration on 400, 2{,}000, and 10{,}000 hours of robot pairs.
Ablations use the 400-hour model and are compared against it (Table~\ref{tab:ablation}; per-task results and two ablations of the 10K-hour model in \cref{tab:ablation_pertask}).

Performance increases with data at every scale.
On \textit{Task Generalization}, which involves no fine-tuning, the average grows fourfold from 400 to 10K hours, against twofold on \textit{Skill Extension}, so pretraining scale matters most when no task data is available.
Without the demonstration, a language-conditioned WAM trained on the same data drops from 24.2\% to 9.0\%, and to 9.5\% on Skill Extension, so the gain comes from the demonstration rather than the data, even after task-specific training.
Removing the wrist views or the state tokens from the demonstration costs 6.8 and 4.9 points, and disabling rollback (Sec.~\ref{sec:align}) costs 3.2 points, as failed steps are no longer redone.

\begin{table}[t]
\centering\small\setlength{\tabcolsep}{3pt}
\caption{\textbf{Data scaling and ablations}, average success rate (\%). Ablations are applied to the 400\,h model. $^{*}$ Without pretraining.}
\label{tab:ablation}
\begin{tabular*}{\columnwidth}{@{\extracolsep{\fill}}lccc@{}}
\toprule
 & \makecell{Task\\Generalization} & \makecell{Skill\\Extension} & All\\
\midrule
0\,h$^{*}$  & --   & 12.3 & --\\
400\,h          & 13.3 & 35.2 & 24.2\\
2{,}000\,h      & 32.3 & 52.8 & 42.6\\
10{,}000\,h     & \textbf{57.5} & \textbf{72.7} & \textbf{65.1}\\
\midrule
w/o demo wrist views  & 10.2 & 24.7 & 17.4\\
w/o demo state        &  9.7 & 29.0 & 19.3\\
w/o rollback          & 11.5 & 30.5 & 21.0\\
w/o demonstration     &  8.5 &  9.5 &  9.0\\
\bottomrule
\end{tabular*}
\end{table}

\subsection{Evaluating on Heterogeneous Demonstrations}
\label{sec:exp_sra}

We next test whether demonstrations from other demonstrators can specify tasks for the executor once translated into its embodiment-grounded representation (Sec.~\ref{sec:representation}) and read through source-routed adapters (Sec.~\ref{sec:plugins}).
We consider five sources: rollouts generated by an LLM agent~\cite{team2026systematically} in simulation, human videos converted by the ego2sim pipeline, UMI recordings with a handheld gripper, and trajectories from two other robots, Astribot S1~\cite{astribot_s1} and Spirit Moz1~\cite{spirit_moz1}.
Both robots are wheeled dual-arm humanoids that differ from our executor in arm kinematics, torso articulation, and camera placement.
For each source, we train one adapter on 200 paired trajectories from 20 tasks disjoint from the evaluation tasks for 500 steps, keeping the pretrained model frozen.
The adapter is then fixed, and the 12 tasks of Sec.~\ref{sec:exp_real} are evaluated under the same protocol, each specified only by a demonstration from that source; w/o SRA feeds the translated demonstration to the frozen model directly.
Demonstrations recorded on the executor pass through no adapter.
\Cref{app:exp_sra} further compares SRAs with full fine-tuning on the same pairs (\cref{tab:sft_vs_sra}) and measures their effect on the predicted actions (\cref{tab:adapter_mae}).

Translation alone already lets the frozen model follow simulated and human demonstrations, and with SRA, demonstrations from all five sources retain 74--91\% of the success obtained with robot demonstrations, although each adapter is trained on only 200 pairs. Simulated and human demonstrations retain the most: both are performed by a simulated copy of the executor, leaving only rendering errors and, for human videos, tracking errors. UMI recordings follow, as they show a handheld gripper rather than the executor, and the two other robots retain the least, as both their appearance and their kinematics differ from the executor's.

\begin{table}[t]
\centering\small\setlength{\tabcolsep}{2.5pt}
\caption{\textbf{Demonstrations from other sources executed on the robot.} Success rate (\%), mean $\pm$ std over the 12 tasks.}
\label{tab:sra}
\begin{tabular*}{\columnwidth}{@{\extracolsep{\fill}}lcc@{}}
\toprule
Demo source & w/o SRA & w/ SRA\\
\midrule
Same-emb. (Galbot G1)     & --              & 65.1 $\pm$ 14.4\\
Sim. (LLM agent)          & 31.8 $\pm$ 16.3 & 59.1 $\pm$ 14.0\\
Human (ego2sim)           & 32.2 $\pm$ 15.4 & 57.9 $\pm$ 14.9\\
UMI                       & 13.1 $\pm$ 7.7  & 55.1 $\pm$ 15.7\\
Cross-emb. (Astribot S1)  & 1.5 $\pm$ 2.6   & 51.0 $\pm$ 16.3\\
Cross-emb. (Spirit Moz1)  & 1.4 $\pm$ 2.2   & 48.2 $\pm$ 14.3\\
\bottomrule
\end{tabular*}
\end{table}

\subsection{Progress Tracking Across Sources}
\label{sec:exp_align}

Since the policy is trained with perturbed anchors (Sec.~\ref{sec:align}), the tracker only needs to place the correct demonstration content inside the window.
We thus measure \emph{window coverage}: annotators mark the matching demonstration time of each execution frame, and we report how often it falls inside the window tracked online as deployed.
We annotate 1{,}000 pairs spanning multiple demonstration sources and tasks (annotation protocol in \cref{app:exp_tracking}).

On same-embodiment pairs, our training-free tracker matches HOST's learned aligner retrained on our robot pairs.
On other sources HOST degrades sharply, to 42.8 on cross-embodiment robots, while ours stays 8--13 points higher.
Wrist views improve coverage on every source, by up to 12.0 points on UMI.
Coverage follows the source order of success in Table~\ref{tab:sra}, suggesting that locating progress is a major part of the remaining cost of a new source.
\Cref{fig:qual} shows a real-world recovery by rollback after a disturbance.

\begin{table}[t]
\centering\small\setlength{\tabcolsep}{3pt}
\caption{\textbf{Window coverage (\%) by demonstration source}; Cross-emb.:\ averages Astribot S1 and Spirit Moz1.}
\label{tab:align}
\begin{tabular*}{\columnwidth}{@{\extracolsep{\fill}}lcccc@{}}
\toprule
Method & \makecell{Same-\\emb.} & Human & UMI & \makecell{Cross-\\emb.}\\
\midrule
HOST~\cite{host}           & 85.5 & 60.2 & 56.2 & 42.8\\
Any2Act, head view only    & 78.4 & 66.5 & 52.1 & 46.9\\
Any2Act                    & \textbf{85.7} & \textbf{73.2} & \textbf{64.1} & \textbf{53.1}\\
\bottomrule
\end{tabular*}
\end{table}

%% file: sec/5_conclusion.tex
\section{Conclusion}
We presented Any2Act, a world action model for robot in-context learning that follows demonstrations from human videos, simulation, UMI recordings, and other robots by translating them into a unified embodiment-grounded representation. Pretrained on robot pairs from 1.2M real trajectories, it performs unseen tasks from a single demonstration; source-routed adapters admit new sources without altering it, and a training-free tracker keeps the context bounded and rolls back to redo failed steps.
Limitations, future directions, and broader impact are discussed in \cref{app:discussion}.

%% file: sec/A_appendix.tex
\twocolumn[{%
  \centering
  {\Large\bfseries Appendix\par}
  \vspace{1.2em}
}]

\setlength{\tabcolsep}{4pt}

\section*{Overview}
\addcontentsline{toc}{section}{Overview}

This appendix complements the main paper in two ways.
First, it specifies \ours completely enough to be reimplemented: the token layout of the model, the progress tracker and its rollback rule, the source-routed adapters, the construction of robot-to-robot pairs, the translation pipeline of every demonstration source, and all hyperparameters (\cref{app:model,app:tracking,app:sra,app:data,app:sources}).
Second, it details the experiments of \cref{sec:exp} (\cref{app:experiments}): the real-world benchmark and baselines, the results of \cref{tab:main} and \cref{tab:ablation} per task, and two further analyses, of how the predicted actions depend on the demonstration and of source-routed adapters against full fine-tuning.
We close with a discussion (\cref{app:discussion}).

\begin{figure*}[t]
\centering
\includegraphics[width=\linewidth]{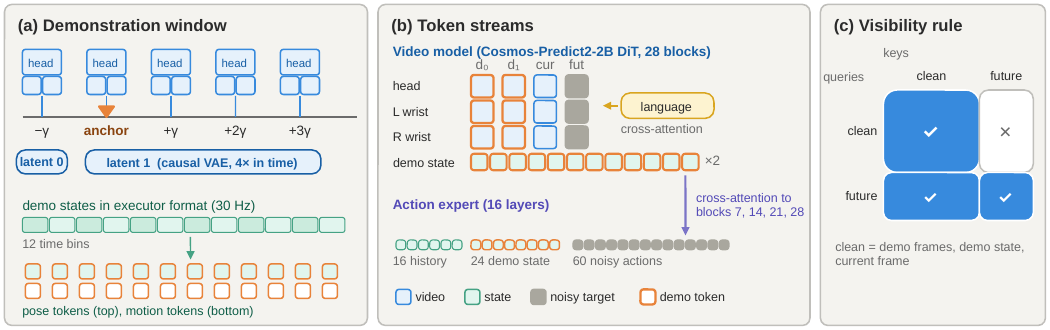}
\caption{\textbf{Inputs and token layout.} (a) The tracked window holds five three-view frames at $\taudem+\gamma\{-1,\dots,3\}$ and the executor-format states between them, which are pooled into 12 pose and 12 motion tokens. (b) The video model reads, per camera, two demonstration latents, the current latent and the noisy future latent, followed by the 24 demonstration-state tokens; the action expert reads the executor history, the same demonstration-state tokens and the noisy action chunk, and cross-attends to four depths of the video model. Orange outlines mark demonstration tokens, the only tokens a source-routed adapter acts on. (c) Clean tokens never attend to the noisy future, so the context is computed once per control step.}
\label{fig:tokens}
\end{figure*}

{\small
\setcounter{tocdepth}{1}
\renewcommand{\contentsname}{\normalsize Contents}
\tableofcontents
}

\section{Model Details}
\label{app:model}

This section specifies what the policy reads and how it is computed; \cref{fig:tokens} summarizes the token layout.

\subsection{Executor Representation}
\label{app:model_rep}

\para{Views.} The executor, a Galbot G1~\cite{galbot_g1}, carries a head camera and two wrist cameras. Head images are resized to $320{\times}320$ and wrist images to $224{\times}224$. A translated demonstration is stored in exactly this format.

\para{State and action space.} Every state and action is a 19-D vector at 30\,Hz: seven joint angles per arm, one opening per gripper (0 closed, 1 open), and the mobile-base pose $(x, y, \psi)$. The base pose is always expressed \emph{relative to the body frame at a reference time} (the current time for the executor, the window start for the demonstration), so that the policy never sees absolute odometry and a demonstration recorded elsewhere in a room is directly comparable.

\para{Prediction.} An action is a chunk of $H{=}60$ absolute states covering the next 2\,s; the model jointly predicts the three camera frames 2\,s ahead, \ie $h{=}60$ in \cref{eq:policy}.

\subsection{Demonstration Window and Tokens}
\label{app:model_window}

\para{Window.} Given an anchor $\taudem$ from the tracker (\cref{app:tracking}), the window samples five images at the nominal times $\taudem + \gamma\cdot\{-1,0,1,2,3\}$\,s, each snapped to the nearest recorded frame, \ie $W_p{=}\gamma$ and $W_f{=}3\gamma$ in \cref{sec:align}, where $\gamma$ is a speed factor. At deployment $\gamma{=}1$; during training $\gamma$ is drawn log-uniformly from $[0.8, 1.25]$ to cover speed differences between demonstrator and executor. Near the start or end of a demonstration the grid is shifted inside the episode instead of being truncated, so the window always holds five real frames when the demonstration is long enough.

\para{Visual tokens.} The five frames of each camera are encoded by the frozen causal Wan VAE of Cosmos-Predict2~\cite{cosmos}, which compresses time by a factor of four and encodes the first frame on its own, so that $1{+}4n$ frames give $1{+}n$ latents. Five frames thus give the two demonstration latents $d_0$ (first frame) and $d_1$ (remaining four) per camera (\cref{fig:tokens}a). Each latent is placed on the temporal axis of the rotary embedding at its time in the window; the current and predicted frames sit at 0 and $+2$\,s.

\para{State tokens.} The demonstrator's 30\,Hz states inside the window, expressed in the executor's state space, (at most 151 rows) are split into 12 equal time bins. Each bin yields a \emph{pose} token (the per-channel mean; grippers take their last value) and a \emph{motion} token (the change over the bin divided by its duration; grippers take their open/close change), giving 24 tokens that convey timing and motion far more densely than the two visual latents. The executor contributes its 16 most recent states.

\subsection{Video Model and Visibility Rule}
\label{app:model_video}

The video model is the 2B-parameter DiT of Cosmos-Predict2~\cite{cosmos} (28 blocks, width 2048). Its sequence concatenates, for each of the three cameras, the patchified latents $[d_0, d_1, \text{current}, \text{future}]$, followed by the 24 demonstration-state tokens (\cref{fig:tokens}b). Learned embeddings mark the role of a token (demonstration or query), its camera, and, for demonstration tokens, their time; state tokens are embedded per hand (seven joints and gripper) and for the base. All demonstration tokens pass through a gated attention block, the \emph{DemoAdapter}, which is part of the pretrained weights $\theta$ and distinct from the source-routed adapters of \cref{app:sra}. The language instruction enters every block through cross-attention to frozen T5 embeddings.

\para{Visibility rule.} Demonstration, demonstration-state and current tokens are \emph{clean}: they attend only to clean tokens; the noisy future attends to all tokens (\cref{fig:tokens}c). Because the clean tokens' inputs and keys do not depend on the denoising step, their activations in the video model are identical across denoising steps, which allows them to be computed once per control step (\cref{app:model_infer}).

\subsection{Action Expert and Objective}
\label{app:model_expert}

The action expert is a 16-layer transformer (width 1536, 12 heads, $\approx$0.5B parameters) over the sequence [16 executor-history tokens, 24 demonstration-state tokens, 60 noisy action tokens]. Every fourth layer cross-attends to the hidden states of the video model after blocks 7, 14, 21 and 28, so it reads both low-level geometry and the model's anticipation of the future. The demonstration-state tokens are shared with the video model and give the expert direct access to the demonstrated motion.

\para{Objective.} Both outputs are trained with rectified flow. For the action chunk $\mathbf{a}$ with noise $\boldsymbol{\epsilon}$ and $t\sim\mathcal{U}[0,1]$, the network predicts the velocity $\boldsymbol{\epsilon}-\mathbf{a}$ from $(1-t)\mathbf{a}+t\boldsymbol{\epsilon}$; the squared error weighs the first second of the chunk by 1, the second by 0.5, and the gripper channels twice. The video term follows Cosmos-Predict2 (log-normal $\sigma$ with $\log\sigma\sim\mathcal{N}(0,1)$ and EDM weighting~\cite{karras2022edm}) and is averaged over the three cameras. The total loss is $\mathcal{L}=2\,\mathcal{L}_{\text{action}}+\mathcal{L}_{\text{video}}$, with the action term normalized over the global batch.

\subsection{Inference and Latency}
\label{app:model_infer}

At each control step the tracker updates the anchor, the window is assembled, and actions and future frames are sampled \emph{jointly}: a Karras schedule ($\sigma\in[0.002, 80]$, $\rho{=}7$) with a second-order RES multistep update for the video and Euler steps for the actions. We use 4 denoising steps. The robot executes the first second (30 rows) of each 2\,s chunk before the next chunk is applied, so the policy re-plans at 1\,Hz while every executed action remains conditioned on a two-second look-ahead.

Three caches follow from the visibility rule. The VAE latents of the demonstration window are kept on the GPU and reused across chunks while the window does not change; the keys and values of the clean tokens (demonstration, demonstration state, current views) are computed once per observation and reused across the denoising steps; the cross-attention keys and values of the instruction are computed once per demonstration.

\Cref{tab:latency} shows how the chunk latency of 180\,ms reported in \cref{sec:model} is reached on one H200 GPU at batch size 1. Each row is the median of 50 timed calls after warm-up, covering preprocessing, encoding of the current views, denoising and the transfer of the actions to the CPU. The tracker update (DINOv3 on the current and previous frame of the three views, DTW update and rollback test) adds 19.4\,ms per chunk. For reference, in the same harness GR00T N1.7 takes 108\,ms per chunk, $\pi_{0.5}$ 176\,ms, HOST 2,142\,ms and Zero-WAM 6,952\,ms.

\begin{table}[t]
\centering
\caption{\textbf{Chunk latency} (60 actions, 2\,s) on one H200 GPU, batch size 1; each row adds to the previous one. Rows above the line use the 10-step sampler.}
\label{tab:latency}
\footnotesize
\begin{tabular}{@{}lrr@{}}
\toprule
Configuration & Median (ms) & P95 (ms) \\
\midrule
FP32 weights, BF16 autocast & 1215.9 & 2058.3 \\
+ demonstration VAE cache & 1195.0 & 2123.1 \\
+ clean-token K/V cache & 1167.0 & 2183.8 \\
+ compact conditioning path, CUDA graph & 730.1 & 883.4 \\
+ BF16 weights & 450.6 & 660.2 \\
+ smaller kernel changes$^\ast$ & 409.7 & 558.4 \\
+ whole sampler in one CUDA graph & 358.4 & 370.4 \\
+ compiled video model and action expert & 220.3 & 231.4 \\
\midrule
+ 4 denoising steps (deployed) & \textbf{180.3} & \textbf{192.7} \\
\bottomrule
\multicolumn{3}{@{}p{0.95\linewidth}@{}}{\scriptsize $^\ast$Plain linear layers in the FFN, text embeddings and text K/V on the GPU, fused QKV projection, batched VAE encoding of the wrist views, output head only for the future tokens.}
\end{tabular}
\end{table}

\section{Progress Tracking and Rollback}
\label{app:tracking}

\begin{figure*}[t]
\centering
\includegraphics[width=\linewidth]{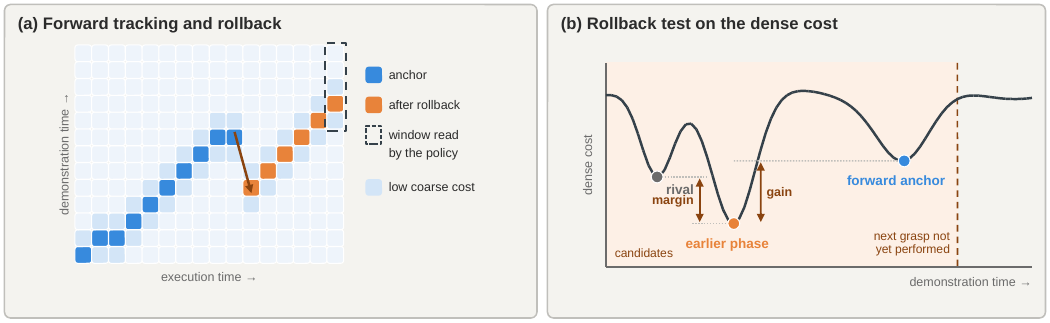}
\caption{\textbf{Progress tracking} (schematic). (a) Coarse cost between execution and demonstration (light: low), the monotone anchor found by online DTW (blue), a rollback after a failed step (arrow), and the anchors that follow (orange). The dashed box is the window the policy reads. (b) The rollback test. Among earlier phases that precede the next grasp the executor has not yet performed (shaded), the minimum of the dense cost must undercut the forward anchor by a \emph{gain} and every rival minimum by a \emph{margin}.}
\label{fig:tracker}
\end{figure*}

\Cref{sec:align} splits alignment in two: a rough, training-free tracker places the current moment inside the window, and the policy, trained to tolerate the tracker's errors, pinpoints its progress within it. We follow the order of the main paper: the deployed tracker (\cref{app:tracking_forward}), the training anchors and their perturbation (\cref{app:tracking_tolerance}), and the recovery by rollback, whose details the main paper defers to this section (\cref{app:rollback}). \Cref{fig:tracker} and \cref{alg:tracker} summarize the procedure.

\subsection{Tracking at Deployment}
\label{app:tracking_forward}

\para{References.} When a demonstration is loaded, its frames are subsampled to $R$ reference times at 1\,Hz. For each reference and each of the three views we store frozen DINOv3 ViT-B/16 features~\cite{simeoni2025dinov3}, both as $4{\times}4$ average-pooled descriptors (coarse) and as patch features (dense).

\para{Coarse cost and update.} At execution step $t$, the coarse cost to reference $s$ is the mean over the three views of $1-\cos$ between pooled descriptors, which is the matching cost $c_t(s)$ of \cref{sec:align} with $w_v{=}1/3$. Since only the past is observed, only the forward part of \cref{eq:align} is computed:
\begin{equation}
F_t(s)=c_t(s)+\min_{0\le\delta\le K}\big[F_{t-1}(s-\delta)+\lambda(\delta)\big],
\label{eq:dtw}
\end{equation}
with $K{=}4$, a stay penalty $\lambda(0){=}0.01$ and a skip penalty $\lambda(\delta){=}0.03(\delta-1)$ for $\delta\ge1$; entries before the current anchor are never revisited, so the deployed anchor $\taudem_t=\arg\min_s F_t(s)$ is monotone, and the window is $[\taudem_t-W_p,\taudem_t+W_f]$ with $W_p{=}1$\,s and $W_f{=}3$\,s (\cref{app:model}). If $c_t(\taudem_t)>0.7$ the match is marked unreliable: the forward anchor is still used, but the rollback test below is skipped.

\subsection{Training Anchors and Perturbation}
\label{app:tracking_tolerance}

\para{Training anchors.} During training the whole query recording is in view, so \cref{eq:align} is evaluated in both directions and the anchor $\tau_t=\arg\min_s[F_t(s)+B_t(s)]$ lies on the globally optimal path. We evaluate it once per pair, offline, when the pair is built (\cref{app:data_pairs}). Since the path is optimized globally rather than estimated greedily from the past, this offline evaluation does not need the step limit and penalties that stabilize the deployed tracker, and its matching cost can also compare the gripper events of the two recordings, which are both known in advance; \cref{app:data_pairs} gives the exact cost.

\para{Perturbation.} The window is anchored at the perturbed anchor $\tau_t+\epsilon$. The deviation $\epsilon$ is drawn from $\mathcal{N}(0, 0.4\,\text{s})$, which spans the typical deviation of the deployed tracker, and truncated to $[-2.9\gamma, 0.9\gamma]$\,s so that $\tau_t$ stays inside the window $[\tau_t+\epsilon-\gamma,\ \tau_t+\epsilon+3\gamma]$. With probability 0.2 it is instead drawn far behind ($[-2.9\gamma,-2.6\gamma]$\,s) or far ahead ($[0.6\gamma,0.9\gamma]$\,s), the cases that a lagging or overshooting tracker produces. The window is randomly stretched or compressed in time by the speed factor $\gamma$ (\cref{app:model}).

\para{Consequence for the tracker.} Because the policy learns to locate its own progress inside the window, the deployed tracker only needs \emph{coverage}: the correct demonstration time must fall inside the window. Since the window extends $3\gamma$ ahead of the anchor and $\gamma$ behind it, it tolerates an anchor that lags the true progress by up to 3\,s but leads it by at most 1\,s. \Cref{tab:align} reports this coverage for every source (\cref{app:exp_tracking}).

\subsection{Rollback}
\label{app:rollback}

The monotone tracker cannot move back by itself, yet a failure often returns the scene to a phase the demonstration has already shown. \Cref{sec:align} therefore re-localizes the anchor to an earlier phase when it explains the current observation clearly better. Below, ``clearly better'' is made precise by a gain and a margin, measured with a cost that is sensitive to the task-relevant part of the scene.

\para{Motion-weighted dense cost.} For each reference we build a patch-weighting that concentrates on what changes in the demonstration. On the head view, 80\% of the weight is spread over the patches whose features move within $\pm2$\,s of the reference (above the 85th percentile, dilated by one patch) and 20\% is uniform. On each wrist view, the weight is proportional to how much the corresponding arm moves within $\pm0.5$\,s; an arm with less than 15\% of the motion is dropped, and the bottom quarter of the wrist image, which shows mostly the gripper itself, is down-weighted to 0.25. The dense cost compares the weighted patch features of the current \emph{and} the 0.1\,s earlier frame with each reference, which suppresses single-frame noise.

\para{Candidates.} The dense cost is minimized over the references from $L_{\text{rb}}$ before the current anchor up to the next grasp of the demonstration that the executor has not yet performed. Grasps are counted from the executor's gripper and re-synchronized with the anchor, so this upper bound always lies after the anchor; its role is to keep phases beyond an unperformed grasp, which may look similar to the present scene, out of the comparison. Let $\tau^\ast$ be the minimizer. A rollback is considered only if $\tau^\ast$ lies between 1\,s and $L_{\text{rb}}$ \emph{before} the previous anchor $\taudem_{t-1}$. The look-back length $L_{\text{rb}}{=}15$\,s bounds how far back a failure may set the task.

\para{Acceptance.} Let $\kappa$ be the dense cost, $\sigma_{\text{loc}}$ the 10--90 percentile spread of $\kappa$ over the recent references, and $\tau_f$ the forward anchor. The rollback is accepted if
\begin{equation}
\begin{gathered}
\underbrace{\frac{\kappa(\tau_f)-\kappa(\tau^\ast)}{\sigma_{\text{loc}}}}_{\text{gain}}\ge 0.15,\qquad
\kappa(\tau^\ast)\le0.7,\\
\underbrace{\frac{\min_{|\tau-\tau^\ast|>1\,\text{s}}\kappa(\tau)-\kappa(\tau^\ast)}{\sigma_{\text{loc}}}}_{\text{margin}}\ge 0.08,
\end{gathered}
\label{eq:rollback}
\end{equation}
\ie the earlier phase must beat the forward anchor (gain), be unambiguous among all candidates more than 1\,s away (margin), and be a valid match. On acceptance the DTW is restarted at $\tau^\ast$ and the window is re-centered there. No past action is replayed: the policy simply reads an earlier part of the demonstration and, because it was trained to locate its own progress inside the window, redoes the step from the current state.

\begin{algorithm}[t]
\caption{Tracker update at execution step $t$}
\label{alg:tracker}
\small
\begin{algorithmic}[1]
\Require references $\{\mathbf{g}_s,\mathbf{G}_s\}_{s=1}^{R}$ (pooled, weighted dense), state $F_{t-1}$, anchor $\taudem_{t-1}$, grasps done $n_g$
\State $c_t(s)\gets$ coarse cost; update $F_t$ by \cref{eq:dtw}; $\tau_f\gets\arg\min_s F_t(s)$
\If{$c_t(\tau_f)>0.7$} \Return $\tau_f$ \Comment{unreliable: no rollback test}\EndIf
\State update $n_g$ from the executor's gripper (re-synced so that $s_{\max}>\taudem_{t-1}$); $s_{\max}\gets$ time of the $(n_g{+}1)$-th demo grasp
\State $\kappa(s)\gets$ dense cost of current and previous frame to $\mathbf{G}_s$
\State $\tau^\ast\gets\arg\min\{\kappa(s): \taudem_{t-1}-L_{\text{rb}}\le s< s_{\max}\}$
\If{$1<\taudem_{t-1}-\tau^\ast\le L_{\text{rb}}$ \textbf{and} \cref{eq:rollback} holds} \State $F_t\gets\delta_{\tau^\ast}$; $n_g\gets$ grasps before $\tau^\ast$; \Return $\tau^\ast$ \EndIf
\State \Return $\tau_f$
\end{algorithmic}
\end{algorithm}

\section{Source-Routed Adapters}
\label{app:sra}

\Cref{sec:plugins} keeps the pretrained parameters $\theta$ frozen and adapts only how the policy interprets the demonstrations of a new source $k$, through low-rank residual branches that are active only on the demonstration tokens of that source (\cref{eq:sra}). This section gives where the branches are placed, the embodiment adapter used for other robots, and how both are trained.

\subsection{Placement and Routing}
\label{app:sra_placement}

Each branch in \cref{eq:sra} has rank 32; we use the scaling $\alpha{=}32$, so $\alpha/r{=}1$ and the update is exactly $m_k B_k A_k h$. In the video model, branches are attached to the query, key, value and output projections of every self-attention and to both linear layers of every feed-forward block; in the action expert, to the same layers of every block, plus a rank-16 branch on the projection of the demonstration-state tokens. Together these form $\phi_k$, with $\approx$47.2M trainable parameters per source, about 1.9\% of the 2.46B-parameter policy.

The routing mask $m_k$ is 1 exactly for the tokens of a demonstration from source $k$: its visual latents and its state tokens, in both the video model and the action expert, as in \cref{sec:plugins}. Current-observation, language, history, future and action tokens never pass through a branch. Cross-attention layers are left untouched, so the action expert reads the adapted demonstration only through the video model's features and its own adapted state tokens.

\para{Exactness.} Because every branch is multiplied by $m_k$, a demonstration recorded on the executor ($m_k{=}0$) produces exactly the same activations with or without the adapter, so following it behaves exactly as before adaptation. \Cref{app:exp_sra} compares this with full fine-tuning on the same pairs.

\subsection{Embodiment Adapter}
\label{app:sra_emb}

For another robot, joint angles do not mean the same thing as the executor's even after reordering into the shared 19-D layout. The embodiment adapter of \cref{sec:plugins} is a residual MLP ($14{\to}128{\to}128{\to}14$, SiLU, last layer zero-initialized; 20K parameters) that maps the 14 arm joints of a routed demonstration toward the executor's joint space before the state tokens are pooled. Being zero-initialized, it starts as the identity. It is trained jointly with the source's SRA on the same pairs and is the only component that differs between robot and non-robot sources.

\subsection{Training}
\label{app:sra_train}

As in \cref{sec:exp_sra}, each adapter is trained for 500 steps on 200 paired trajectories from 20 tasks that are disjoint from the 12 evaluation tasks, with $\theta$ frozen: 10 source demonstrations per task, each paired with one executor recording of the same task in a different scene. In every pair the \emph{query} is an executor trajectory and the \emph{demonstration} comes from source $k$, so the supervision (actions and future frames) is always the executor's own. Training anchors $\tau_t$ are obtained with \cref{eq:align} offline, as for pretraining pairs (\cref{app:data_pairs}). Because cross-source pairs are already noisier than same-source ones, the perturbation of the anchor is milder ($\epsilon\sim\mathcal{N}(0, 0.2\,\text{s})$, truncated to $\pm0.5$\,s, same range of $\gamma$). Language is never dropped and tasks are sampled uniformly. The remaining hyperparameters are in \cref{tab:hparams}.

\section{Pretraining Data and Recipe}
\label{app:data}

\subsection{Corpus}
\label{app:data_corpus}

The corpus consists of 1.2M teleoperated trajectories (10K hours, 30\,s on average) recorded on Galbot G1 robots during routine data collection, in various environments (homes, offices, retail stores, kitchens and laboratories). Each recording is labeled by a VLM (Qwen3.8~\cite{qwen38}) with a canonical instruction and a task identity. Recordings that share a task identity form a task group; the 10K groups follow a long-tailed size distribution and the task-type composition of \cref{fig:data}. No recording of the twelve evaluation tasks, in any scene, is part of the corpus (\cref{app:exp_benchmark}).

\subsection{Pair Construction}
\label{app:data_pairs}

\begin{figure*}[t]
\centering
\includegraphics[width=\linewidth]{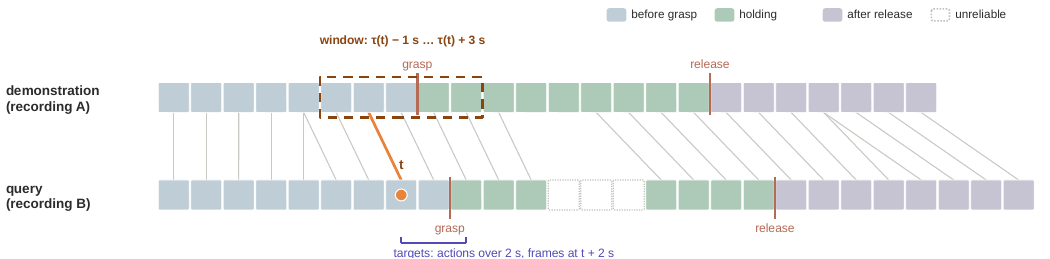}
\caption{\textbf{From recordings to demonstration$\to$query pairs.} Two recordings of the same task are aligned once, offline; at training time a query time is sampled and the matched demonstration time defines the (perturbed) window. $\tau(t)$ is the training anchor $\tau_t$ of \cref{eq:align} at query time $t$, not the deployed anchor $\taudem_t$. Schematic.}
\label{fig:pairs}
\end{figure*}

\Cref{fig:pairs} shows how recordings become training pairs.

\para{Candidates.} Two recordings form a candidate pair if they share task group and have the same \emph{gripper signature}: the same initial gripper states and the same ordered sequence of grasp and release events (interleaved over both arms). The signature removes pairs in which the two operators solved the task with a different strategy (\eg a different arm), whose alignment would be ill-defined.

\para{Alignment.} Each pair is aligned with \cref{eq:align}, evaluated offline over the full query recording, which plays the role of the execution, against the partner recording, which plays the role of the demonstration. Frames are described by the $4{\times}4$ pooled DINOv3 features of the three views, mixed as $0.8\times$ appearance and $0.2\times$ change relative to the episode median, which emphasizes what moves. The matching cost $c_t(s)$ is the cosine distance between these descriptors plus $2\times$ the difference in \emph{stage}, the number of gripper events so far in each recording, which strongly penalizes matching a frame before a grasp to a frame after it. Because the whole path is optimized at once, the advance is not limited ($K$ unbounded, $\lambda\equiv0$). For every query frame $t$, $\tau_t=\arg\min_s[F_t(s)+B_t(s)]$ is the training anchor (\cref{app:tracking_tolerance}).

\para{Quality check.} A query (demonstration) frame is \emph{reliable} if at least 75\% of the frames it is matched to have an appearance distance $\le0.25$ and the same stage. Only reliable intervals are used as query times or matched demonstration times, and a pair is dropped entirely if more than 85\% of its path steps advance only one of the two recordings, which indicates that one recording stalls while the other progresses.

\para{Sampling.} A training sample draws a reliable query time uniformly over the corpus, one of its partners, and a window around the matched demonstration time with the perturbations of \cref{app:tracking_tolerance}. Language is replaced by the empty string with probability 0.2.

\subsection{Recipe and Hyperparameters}
\label{app:data_recipe}

\Cref{tab:hparams} lists all hyperparameters. The pretrained DiT and the newly initialized modules (action expert, embeddings, DemoAdapter) use separate learning rates; training uses FSDP in bf16 with activation checkpointing.

\begin{table}[t]
\centering
\caption{\textbf{Hyperparameters.} Values follow our configuration files.}
\label{tab:hparams}
\scriptsize
\setlength{\tabcolsep}{3pt}
\begin{tabular}{@{}lll@{}}
\toprule
& Pretraining & Source-routed adapter \\
\midrule
Trainable & all (DiT + new modules) & routed LoRA (+ emb.\ adapter) \\
Pairs & same source, robot--robot & robot query, source-$k$ demo \\
Sampling & uniform over query times & task-balanced \\
Optimizer & AdamW (0.9, 0.999), wd 0.01 & AdamW, wd 0 \\
Peak LR & DiT $2{\times}10^{-5}$, new $5{\times}10^{-5}$ & $5{\times}10^{-5}$ \\
Schedule & warm-up 2K, hold to 20K, & warm-up 100, cosine \\
 & linear decay to 25K, then & to 10\% of peak \\
 & constant ($5{\times}10^{-6}$ / $10^{-5}$) & \\
Batch & 768 ($32\times24$) & 192 ($3\times64$) \\
Steps & 200K & 500 \\
Window anchor & $\mathcal{N}(0,0.4\,\text{s})$; 20\% far offsets & $\mathcal{N}(0,0.2\,\text{s})$, $|\cdot|\le0.5$\,s \\
Speed factor $\gamma$ & logU[0.8, 1.25] & logU[0.8, 1.25] \\
Language dropout & 0.2 & 0 \\
LoRA rank / $\alpha$ & -- & 32 / 32 (state proj.\ 16) \\
Loss & $2\,\mathcal{L}_{\text{action}}+\mathcal{L}_{\text{video}}$ & same \\
\bottomrule
\end{tabular}
\end{table}

\section{Translating Demonstrations}
\label{app:sources}

Every source is translated into the unified representation of \cref{sec:representation}: a head view and two wrist views in the executor's camera layout and a 19-D executor state trajectory. An overview is given in \cref{fig:data}. The translations differ in what they must recover: simulation and human videos lack the executor's actions and wrist views, UMI recordings lack joint-space actions, and other robots have the views but a different body. What remains after translation, the \emph{residual gap}, is what the source-routed adapter absorbs.

\begin{figure*}[t]
\centering
\includegraphics[width=\linewidth]{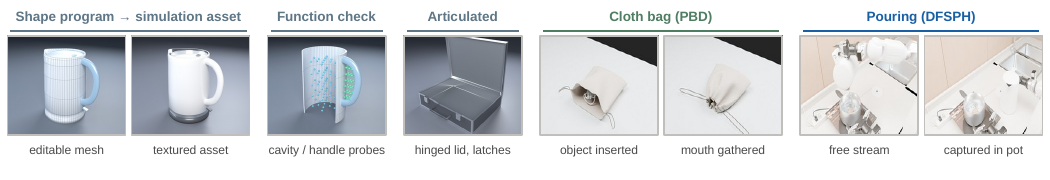}
\caption{\textbf{Assets and physics} shared by both simulation-based sources. Rigid and articulated assets are compiled from editable shape programs and checked for usable openings and grasp channels; bags and liquids are simulated in Genesis.}
\label{fig:assets}
\vspace{10pt}

\includegraphics[width=\linewidth]{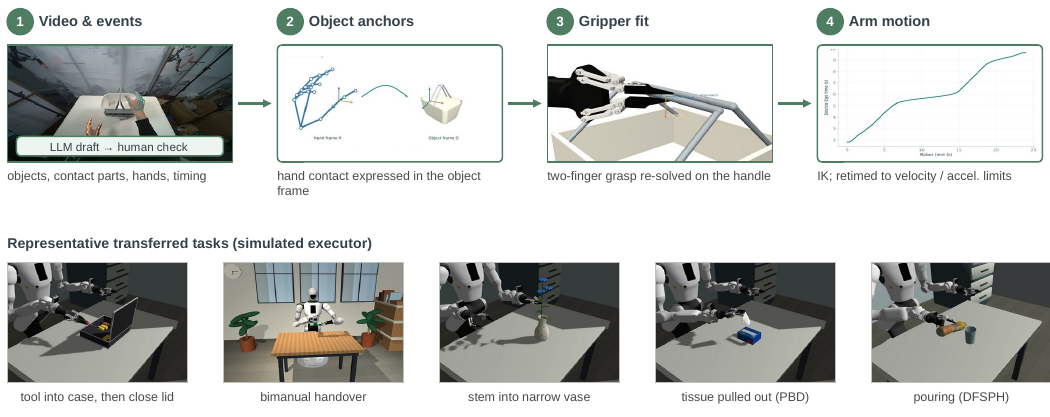}
\caption{\textbf{Human-video source (ego2sim).} Top: what the human does to each object is extracted from the video (steps 1--2) and re-solved for the robot's gripper and arm (steps 3--4). Bottom: transferred tasks on the simulated executor.}
\label{fig:ego2sim}
\vspace{10pt}

\includegraphics[width=\linewidth]{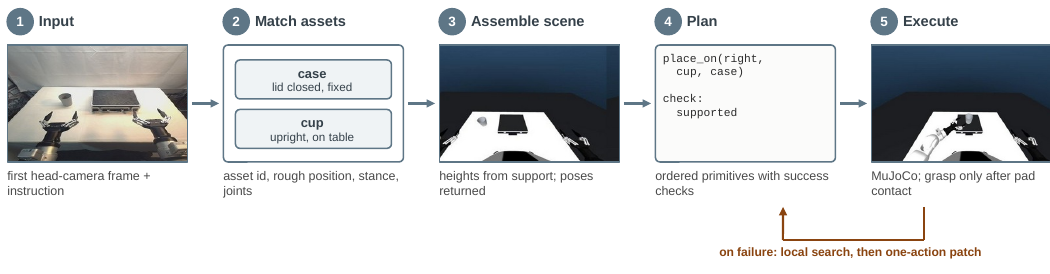}
\caption{\textbf{Simulation source (LLM agent)}, used when no human video is available, for the instruction ``put the cup on the case''. One head-camera frame and the instruction are turned into a simulated demonstration with two LLM calls (steps 2 and 4); heights and contacts are computed by the simulator and trajectories by a local motion planner.}
\label{fig:sim}
\end{figure*}

\para{Overview.} \Cref{app:sources_assets} describes the asset library and physics shared by the two simulation-based sources. \Cref{app:sources_human} converts human videos with ego2sim, and \cref{app:sources_sim} generates demonstrations with an LLM agent when no human video is available; both pipelines are illustrated in \cref{fig:assets,fig:ego2sim,fig:sim}. \Cref{app:sources_umi,app:sources_robots} map UMI recordings and other robots, whose views are kept and only whose motion is translated.

\para{Two sources, one simulator.} Human videos and simulation, (2) and (1) in \cref{sec:representation}, are translated through the same real-to-sim backbone. The scene is rebuilt in a physics simulator from a head-camera image or video and a shared asset library (\cref{app:sources_assets}), and the task is then executed on a simulated executor, a model of the Galbot G1, which directly yields the actions and the three views in the executor's layout. The library also contains models of other robots, so the same rollout can be produced on another body. The two sources differ only in \emph{what specifies the task}. In ego2sim, a person has performed the task in an egocentric video, and the pipeline transfers what the hand does to the objects onto the robot (\cref{app:sources_human}). When no human video of the task is available, the instruction alone specifies it: the LLM agent GPT-6 Astra~\cite{team2026systematically} plans the task in the rebuilt scene and rolls it out on the simulated executor (\cref{app:sources_sim}). ego2sim thus turns abundant human videos, which record how a task is actually done, into demonstrations the executor can follow, and the LLM agent covers tasks for which no such video exists, at no recording cost.

\subsection{Assets and Physics}
\label{app:sources_assets}

The asset library (\cref{fig:assets}) contains hundreds of registered assets (objects, robots and fixtures), many of them instances of shared parametric templates. Frequently used classes can be called by name, with resting poses, joint definitions, annotated grasp candidates and operation points on lids, doors and knobs.

\para{Rigid and articulated assets.} An asset is built from calibrated robot views of the real object. An LLM describes its parts, joints and openings, citing the frames that show them and marking what cannot be seen (\eg wall thickness) as unknown. The description is compiled into a short shape program (revolved, lofted and swept profiles, boxes and hinges) that produces the visual and the collision mesh from the same parameters. The program is then scaled to metric size using the calibrated cameras, silhouettes and the table plane. Before use, we check each asset for manipulation: openings must be clear, cavities connected, handles must leave room for the fingers (probe points in \cref{fig:assets}), and joints must have the observed range. A person then inspects the asset and refines it where needed.

\para{Bags and liquids.} Cloth, paper and open bags are triangle shells simulated with position-based dynamics (PBD) in Genesis, and filled pouches use implicit FEM. For pouring, the source container holds a finite amount of liquid; when it is tilted over the target, particles are emitted at the spout and their free flow is simulated with DFSPH. Material parameters (stiffness, density, viscosity) are nominal values, not measurements.

\subsection{Human Videos via ego2sim}
\label{app:sources_human}

Our agentic ego2sim builds on EgoEngine~\cite{liu2026egoengine}, which reconstructs egocentric hand--object interaction in simulation. As in \cref{sec:representation}, it estimates hand and object poses from the video, rebuilds the scene, transfers the grasp to the robot gripper and executes the human trajectory on the simulated executor (\cref{fig:ego2sim}). A human hand and a two-finger gripper cannot follow the same path, so the trajectory is transferred at the level of the objects: ego2sim keeps what the human does \emph{to the objects}, namely which object is touched, at which part, by which hand, in which order and to which end state, and solves again how the robot achieves it. For example, when a person grasps a screwdriver by its handle, puts it into a case and closes the lid, ego2sim keeps three facts: the contact is on the handle, the tool ends up fully inside the case, and the lid is then closed. The grasp, the path over the case wall and the arm pose for pushing the lid are solved for the robot.

\para{(1) Hand poses and events.} Hand keypoints are tracked in the video, and an LLM drafts the events: objects and their roles, the contact part, the acting hand, grasp and release times, and the target relation (``inside'', ``on'', ``lid closed''). A reviewer checks the draft against the video, especially contacts hidden by the hand, which takes 3.5\,min per video on average. Ordering constraints are kept explicitly: in a direct handover the receiving hand must hold the object before the giving hand releases it, whereas a handover via the table requires the object to rest on the table in between.

\para{(2) Object poses and scene.} The objects are matched to library assets, or reconstructed (\cref{app:sources_assets}), and registered in the scene with the calibrated cameras and the reviewed table layout. Each contact is stored in the frame of the object, or of the moving part such as a lid, so it moves with the object and stays valid if the object starts at a different pose.

\para{(3) Grasp transfer.} Around each contact anchor we sample grasps (approach direction, rotation, opening) and solve the seven arm joints and the gripper opening by bounded least squares, so that both finger pads land on opposite faces of the object without other collisions. A grasp is accepted only if the same grasp also works for the rest of the carry (lift, transfer, place) on one continuous IK branch; a grasp that is reachable on the table but hits the case wall later is replaced by another candidate.

\para{(4) Execution.} While an object is held, its target poses (from the reviewed end relation and waypoints) are converted into gripper poses through the fixed grasp transform. IK is solved frame by frame, starting from the previous solution; the redundant degree of freedom of the 7-DoF arm is used to avoid self-collision, and in narrow spaces a joint-space planner (RRT-Connect) takes over. The motion is retimed to the robot's velocity and acceleration limits, and the mapping between robot time and video time is kept (\cref{fig:ego2sim}, step 4). Rigid objects are attached to the gripper only after both pads pass a contact check, and fall under free dynamics after release; bags and liquids are simulated as in \cref{app:sources_assets}, with the robot and the held container as moving boundaries.

A translated demonstration is accepted if the arm motion is valid (joint limits, pad contact, clearance) and the task outcome matches the video (end relations and event order). Human and UMI demonstrators are static in our data, so their base channels are zero (\cref{app:model_rep}).

\subsection{Simulation via an LLM Agent}
\label{app:sources_sim}

When no human video of a task is available, a demonstration can still be produced from the instruction alone. Following recent real-to-sim pipelines~\cite{zhao2025robot,li2026arstag}, we model the current demonstration scene from one head-camera frame, and the LLM agent GPT-6 Astra~\cite{team2026systematically} rolls out the task on the simulated executor, generating its actions directly, as in \cref{sec:representation} (\cref{fig:sim}). The agent is called twice, once to recognize the scene and once to plan. It gives only a rough layout and an ordered plan; heights and contacts are computed by the simulator, and trajectories by a local motion planner~\cite{wang2024genh2r,wang2025mobileh2r}, which also lets failed attempts be retried freely.

\para{(1) Input.} A head-camera frame of the current demonstration scene and the instruction.

\para{(2) Matching.} The first call receives the frame, the instruction and the catalog of callable assets (\cref{app:sources_assets}). For every visible object it returns the asset id, an approximate position on the table, the yaw, a \emph{discrete} resting pose, the initial joint state, \eg an open case lid, and support relations such as ``on the case''. The resting pose matters: a pack of wipes standing on its side cannot be obtained from a lying one by a yaw rotation.

\para{(3) Assembly.} The scene is built in MuJoCo~\cite{mujoco}. Heights follow from collision geometry and support relations, so an object on top of the case is placed on the case's top surface. The resulting poses, hinge axes and annotated grasp and operation points are passed on to the planner.

\para{(4) Planning.} The second call receives these actual coordinates and writes an ordered list of primitives (\texttt{place\_in}, \texttt{place\_on}, \texttt{open\_lid}, \texttt{close\_lid}, \texttt{open\_door}, \texttt{close\_door}, \texttt{pour}, \texttt{handover}, \texttt{brace}) for the two arms of the executor, each with an arm, an object, a target and, where needed, a sub-goal such as ``fully inside'' before a lid is closed. Each primitive comes with its own success checks (contained, supported, upright, joint angle).

\para{(5) Execution.} A local controller picks a feasible annotated grasp, computes a collision-free joint trajectory for the simulated executor and executes it in MuJoCo. A grasp holds the object only after both finger pads have touched it from opposite sides for 30\,ms, so the simulator cannot carry an object the gripper did not reach. When a primitive fails, the controller first tries other grasps or trajectories on a copy of the current state, and only then is the planner asked to replace the failed primitive, without changing the other primitives or the goal. A verified demonstration takes about 1.2\,min.

\subsection{UMI}
\label{app:sources_umi}

UMI recordings~\cite{umi} already contain a head view and two wrist views, but their motion is a 6-DoF gripper pose and a gripper width. We express the poses in the executor's base frame with a fixed calibration between the UMI head rig and the executor's head camera, solve IK on the executor's arms with the solver of \cref{app:sources_human}, and map the width linearly to the executor's gripper opening. The views are kept as recorded; the handheld gripper and the human arm remain visible, which is the residual gap the adapter must absorb.

\subsection{Other Robots}
\label{app:sources_robots}

Astribot S1~\cite{astribot_s1} and Spirit Moz1~\cite{spirit_moz1} are wheeled dual-arm humanoids with a head camera and two wrist cameras. Their views are used as recorded. Their joint states are re-ordered into the shared layout (seven joints per arm, one opening per gripper, base $x,y,\psi$), and the base is expressed relative to the body frame as for the executor. Arm kinematics, torso articulation and camera placement still differ, which is why the embodiment adapter (\cref{app:sra}) is trained together with these sources' SRAs.
\section{Experiments}
\label{app:experiments}

This section follows \cref{sec:exp}. We first define the real-world benchmark and the baselines (\cref{app:exp_benchmark,app:exp_baselines}) and give the results of \cref{tab:main} per task (\cref{app:exp_pertask}). We then show how the predicted actions and the executed behavior depend on the demonstration (\cref{app:exp_demo}), give the RoboTwin 2.0 protocol (\cref{app:exp_robotwin}), and expand scaling and ablations per task, including two ablations of the full model (\cref{app:exp_ablation}). Finally, we turn to the source-routed adapters that admit other demonstrators (\cref{app:exp_sra}) and to progress tracking and recovery (\cref{app:exp_tracking}).

\subsection{Real-World Benchmark}
\label{app:exp_benchmark}

\para{Tasks.} \Cref{tab:tasks} describes the twelve tasks of \cref{fig:exp}. Task Generalization composes skills seen in pretraining into new goals, step orders and object combinations; Skill Extension requires skills beyond pretraining, such as handling a cloth bag, an articulated cabinet or bimanual pouring. A task is long-horizon ($\dagger$) if its demonstration has at least five steps.

\para{Properties.} The twelve tasks were defined after the pretraining corpus was frozen, and no recording of them, in any scene, is part of it. For each task we also retrieved the 20 corpus task groups with the most similar instruction (Qwen3-Embedding-8B) and inspected them by hand: none has the same goal and object combination.

\para{Trials.} Each task is evaluated in 100 trials per method, split over five demonstrations and five test scenes. The demonstrations were recorded once, before any policy was run, in scenes different from all test scenes. Within a scene, object instances and placements are re-randomized from a fixed list for every trial, the same for all methods, and methods are evaluated in interleaved blocks of ten trials. An operator scores each trial with the criteria of \cref{tab:tasks}, within a time limit of twice the demonstration duration.

\begin{table*}[t]
\centering
\caption{\textbf{Real-world tasks.} Steps: annotated subtasks of the demonstration. $\dagger$: long-horizon.}
\label{tab:tasks}
\scriptsize
\setlength{\tabcolsep}{3pt}
\renewcommand{\arraystretch}{1.0}
\renewcommand{\tabularxcolumn}[1]{m{#1}}
\begin{tabularx}{\linewidth}{@{}m{0.85cm} m{1.35cm} X c X@{}}
\toprule
& Task & Task description & Steps & Success criterion \\
\midrule
\raisebox{-0.0\height}{\includegraphics[width=0.8cm,page=1]{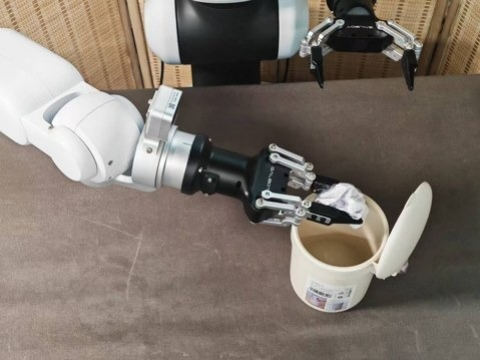}}
& Discarding & Discard the trash scattered on the table into the trash bin. & 2
& All target trash is inside the bin. \\
\raisebox{-0.0\height}{\includegraphics[width=0.8cm,page=2]{fig/tasks.pdf}}
& Lidding & Cover the pot with its lid. & 2
& The lid is properly seated on the pot. \\
\raisebox{-0.0\height}{\includegraphics[width=0.8cm,page=3]{fig/tasks.pdf}}
& Gathering & Place multiple fruits and vegetables into the basket. & 4
& All target fruits and vegetables are inside the basket. \\
\raisebox{-0.0\height}{\includegraphics[width=0.8cm,page=4]{fig/tasks.pdf}}
& Pinching & Put away the napkins and fruit. & 4
& The napkins and fruit are in their designated storage locations. \\
\raisebox{-0.0\height}{\includegraphics[width=0.8cm,page=5]{fig/tasks.pdf}}
& Filling$^\dagger$ & Place the vegetables into the stainless-steel pot, then cover it with the lid. & 5
& All target vegetables are inside the pot, and the lid is properly seated. \\
\raisebox{-0.0\height}{\includegraphics[width=0.8cm,page=6]{fig/tasks.pdf}}
& Packing$^\dagger$ & Place the potato chip canisters on the serving plate one by one, then move the plate. & 6
& The plate reaches the target location with all canisters remaining on it. \\
\midrule
\raisebox{-0.0\height}{\includegraphics[width=0.8cm,page=7]{fig/tasks.pdf}}
& Relaying$^\dagger$ & Place the corn into the steaming bowl, place the bowl inside the pot, then cover the pot. & 6
& The corn is inside the bowl, the bowl is inside the pot, and the lid is properly seated. \\
\raisebox{-0.0\height}{\includegraphics[width=0.8cm,page=8]{fig/tasks.pdf}}
& Sorting$^\dagger$ & Put away assorted items and use the watering can. & 6
& All target items are in their designated storage locations, and the watering action is completed. \\
\raisebox{-0.0\height}{\includegraphics[width=0.8cm,page=9]{fig/tasks.pdf}}
& Storing$^\dagger$ & Place the shoes into the shoe cabinet one by one, then close the cabinet door. & 8
& All target shoes are inside the cabinet, and the door is closed. \\
\raisebox{-0.0\height}{\includegraphics[width=0.8cm,page=10]{fig/tasks.pdf}}
& Bagging & Place the hair dryer into the cloth bag. & 4
& The hair dryer is fully inside the cloth bag. \\
\raisebox{-0.0\height}{\includegraphics[width=0.8cm,page=11]{fig/tasks.pdf}}
& Extracting$^\dagger$ & Remove the books from the bookshelf one by one and pack them into the box. & 6
& All target books have been transferred from the bookshelf into the box. \\
\raisebox{-0.0\height}{\includegraphics[width=0.8cm,page=12]{fig/tasks.pdf}}
& Pouring$^\dagger$ & Pour water into the mug through coordinated use of both arms. & 6
& Water is successfully transferred into the mug. \\
\bottomrule
\end{tabularx}
\end{table*}

\subsection{Baselines}
\label{app:exp_baselines}

\para{VLAs.} $\pi_{0.5}$~\cite{pi05} and GR00T N1.7~\cite{gr00t_n17} are fine-tuned from their public checkpoints on 50 teleoperated demonstrations of each evaluation task. With 20 demonstrations per task, the number given to the in-context policies on Skill Extension, they reach only 19.2\% and 23.5\% on average, so we report the 50-demonstration setting.

\para{In-context policies.} HOST~\cite{host} and Zero-WAM~\cite{zerowam} are retrained with their official architectures and recipes on our robot pairs; HOST's learned aligner is trained on our DTW alignments, and Zero-WAM receives the full demonstration as a prefix. On Skill Extension they are adapted with the same 20 demonstrations per task as \ours.

\subsection{Real-World Results per Task}
\label{app:exp_pertask}

\Cref{tab:main_pertask} repeats \cref{tab:main} with 95\% Wilson intervals for \ours. Over the 600 trials of each benchmark, \ours exceeds the strongest baseline, GR00T N1.7, by 10.0 points on Task Generalization and by 25.2 points on Skill Extension ($p<0.001$ for both, one-sided Fisher exact test). The largest margins are on Gathering, Sorting and Pouring, where the demonstration conveys the order of the steps and the coordination of the two arms.

\begin{table}[t]
\centering
\caption{\textbf{Real-world success (\%) per task}, as in \cref{tab:main}, with 95\% Wilson intervals for \ours. Bold: best per row. $\dagger$: long-horizon.}
\label{tab:main_pertask}
\scriptsize
\setlength{\tabcolsep}{2.4pt}
\begin{tabular}{@{}ll cccc c@{}}
\toprule
 & Task & $\pi_{0.5}$ & GR00T N1.7 & HOST & Zero-WAM & \textbf{Any2Act} [95\% CI] \\
\midrule
\multirow{6}{*}{\rotatebox{90}{\scriptsize Task Gen.}}
 & Discarding & 37 & 62 & 19 & 24 & \textbf{64} \,{\scriptsize[54,\,73]} \\
 & Lidding & 41 & 33 & 26 & 16 & \textbf{47} \,{\scriptsize[38,\,57]} \\
 & Gathering & 43 & 57 & 0 & 17 & \textbf{77} \,{\scriptsize[68,\,84]} \\
 & Pinching & 24 & 28 & 13 & 19 & \textbf{38} \,{\scriptsize[29,\,48]} \\
 & Filling$^\dagger$ & \textbf{57} & 44 & 11 & 8 & 56 \,{\scriptsize[46,\,65]} \\
 & Packing$^\dagger$ & 56 & 61 & 25 & 34 & \textbf{63} \,{\scriptsize[53,\,72]} \\
\midrule
\multirow{6}{*}{\rotatebox{90}{\scriptsize Skill Ext.}}
 & Relaying$^\dagger$ & \textbf{63} & 42 & 13 & 21 & 61 \,{\scriptsize[51,\,70]} \\
 & Sorting$^\dagger$ & 19 & 39 & 33 & 46 & \textbf{88} \,{\scriptsize[80,\,93]} \\
 & Storing$^\dagger$ & 48 & 67 & 35 & 65 & \textbf{79} \,{\scriptsize[70,\,86]} \\
 & Bagging & 29 & 45 & 13 & 7 & \textbf{55} \,{\scriptsize[45,\,64]} \\
 & Extracting$^\dagger$ & 61 & 69 & 31 & 33 & \textbf{84} \,{\scriptsize[76,\,90]} \\
 & Pouring$^\dagger$ & 39 & 23 & 25 & 27 & \textbf{69} \,{\scriptsize[59,\,77]} \\
\midrule
\multicolumn{2}{@{}l}{Task Gen. avg} & 43.0 & 47.5 & 15.7 & 19.7 & \textbf{57.5} \,{\scriptsize[53.5,\,61.4]} \\
\multicolumn{2}{@{}l}{Skill Ext. avg} & 43.2 & 47.5 & 25.0 & 33.2 & \textbf{72.7} \,{\scriptsize[69.0,\,76.1]} \\
\multicolumn{2}{@{}l}{All} & 43.1 & 47.5 & 20.3 & 26.4 & \textbf{65.1} \,{\scriptsize[62.3,\,67.7]} \\
\bottomrule
\end{tabular}
\end{table}

\subsection{Sensitivity to the Demonstration}
\label{app:exp_demo}

An in-context policy should change its behavior when the demonstration shows something different, but not when another demonstration of the same task is given. We test this offline on the predicted actions and on the real robot.

\para{Action error.} On the six Task Generalization tasks, which share one set of weights, we take 50 query chunks (2\,s, 14 arm joints) per task from 10 executor recordings of the task, disjoint from its demonstrations, and predict every chunk with each of the five demonstrations of the task (\emph{same task}) and with the 25 demonstrations of the other five tasks (\emph{other task}); the window is placed by the deployed tracker. \Cref{tab:demo_mae} shows that the five demonstrations of a task lead to nearly the same error (32.0--35.7$\times10^{-3}$\,rad on average), whereas a demonstration of another task raises it by 20.1$\times10^{-3}$\,rad (59\%), with every interval above zero.

\para{Correct \vs wrong demonstrations.} On the real robot, we give the policy either a correct demonstration of the instructed task or a wrong one that conflicts with the instruction, \eg by showing another object or another step order. We use six tasks: two pick-and-place tasks, two long-horizon placing tasks whose step order is easily confused, book extraction and pouring, with 50 trials per task and condition. With the correct demonstration (\cref{tab:demo_swap}), the robot keeps the instructed order in 96.3\% of the trials and succeeds in 67.3\%. With a wrong one, success falls to 23.7\%, and on the two order-sensitive tasks only 14 and 32\% of the trials keep the instructed order. Both offline and on the robot, the policy executes what the demonstration shows, not only what the instruction or the current scene suggests.

\begin{table}[t]
\centering
\caption{\textbf{Action error with same- and other-task demonstrations} ($10^{-3}$\,rad, 14 arm joints; 50 chunks from 10 executor recordings per task). Same task: mean and range over the five demonstrations of the task. Other task: mean over the 25 demonstrations of the other tasks. $\Delta$: difference, with 95\% bootstrap interval over query recordings.}
\label{tab:demo_mae}
\footnotesize
\setlength{\tabcolsep}{3pt}
\begin{tabular}{@{}lcccc@{}}
\toprule
Task & Same task & Range & Other task & $\Delta$ [95\% CI] \\
\midrule
Discarding & 22.6 & 22.2--23.2 & 30.2 & +7.6 [+5.3, +10.1] \\
Lidding & 25.1 & 23.7--26.3 & 67.2 & +42.1 [+37.1, +46.8] \\
Gathering & 32.1 & 30.4--33.1 & 42.3 & +10.2 [+7.7, +12.8] \\
Pinching & 25.3 & 24.7--25.8 & 43.0 & +17.7 [+14.3, +21.3] \\
Filling$^\dagger$ & 36.8 & 34.9--37.9 & 67.2 & +30.4 [+19.4, +43.5] \\
Packing$^\dagger$ & 62.1 & 56.2--68.1 & 74.9 & +12.9 [+3.8, +21.3] \\
\midrule
Average & 34.0 & 32.0--35.7 & 54.1 & +20.1 [+17.4, +23.0] \\
\bottomrule
\end{tabular}
\end{table}

\begin{table}[t]
\centering
\caption{\textbf{Correct \vs wrong demonstrations on the real robot} (50 trials per task and condition). Succ.: success on the instructed task (\%). Order: trials whose steps follow the instructed order (\%).}
\label{tab:demo_swap}
\footnotesize
\setlength{\tabcolsep}{3pt}
\begin{tabular}{@{}lcccc@{}}
\toprule
& \multicolumn{2}{c}{Correct demo} & \multicolumn{2}{c}{Wrong demo} \\
\cmidrule(lr){2-3}\cmidrule(lr){4-5}
Task & Succ. & Order & Succ. & Order \\
\midrule
Pick tissue box & 78 & 100 & 14 & 100 \\
Pick chip canister & 66 & 100 & 42 & \phantom{0}80 \\
Apple, pumpkin, gourd $\to$ pot & 86 & \phantom{0}96 & 24 & \phantom{0}14 \\
Vase, cup $\to$ storage box & 74 & 100 & 40 & \phantom{0}32 \\
Extract books into box & 52 & \phantom{0}90 & \phantom{0}8 & \phantom{0}22 \\
Pour water & 48 & \phantom{0}92 & 14 & \phantom{0}12 \\
\midrule
Average & 67.3 & 96.3 & 23.7 & 43.3 \\
\bottomrule
\end{tabular}
\end{table}

\subsection{RoboTwin 2.0}
\label{app:exp_robotwin}

We follow the cross-task protocol of Zero-WAM~\cite{zerowam} on RoboTwin 2.0~\cite{robotwin2}. Starting from the 10K-hour checkpoint, \ours is post-trained on demonstration$\to$query pairs of the 43 seen tasks (50 demonstrations per task, 20K steps) and evaluated on the 7 unseen tasks with three seeds of 100 rollouts. Each rollout receives as its demonstration one simulated execution of the unseen task, recorded with a different seed and rendered from RoboTwin's head and wrist cameras, with its joint states. No adapter is used, since simulation is the post-training domain. Baseline numbers are taken from~\cite{zerowam}.

\subsection{Scaling and Ablations per Task}
\label{app:exp_ablation}

\Cref{tab:ablation_pertask} expands \cref{tab:ablation} per task and adds two ablations of the full 10K-hour model. The gain from scale is not carried by a few tasks: from 400 to 10K hours every Task Generalization task improves by 38 to 54 points, and this zero-shot benchmark grows $4.3\times$, against $2.1\times$ for Skill Extension. At 400 hours, removing the demonstration hurts most on three long-horizon Skill Extension tasks, Relaying (40 to 0), Storing (39 to 0) and Extracting (54 to 5), even though the language-conditioned model is fine-tuned on the same 20 demonstrations per task: there the demonstration carries what the fine-tuning data does not, the order of the steps.

Both components matter more as the model scales. At 10K hours, removing the demonstration lowers success from 65.1 to 26.2\%; on Task Generalization the gap to the language-conditioned model grows from 4.8 points at 400 hours to 43.8 points, so scale benefits a policy that reads demonstrations far more than one that reads only language. Disabling rollback costs 11.0 points at 10K hours, against 3.2 points at 400 hours (\cref{tab:ablation}), most on Extracting (84 to 46), Packing (63 to 42) and Bagging (55 to 35).

\begin{table*}[t]
\centering
\caption{\textbf{Scaling and ablations per task} (success, \%), expanding \cref{tab:ablation} and adding two ablations of the 10,000-hour model. $\dagger$: long-horizon.}
\label{tab:ablation_pertask}
\small
\setlength{\tabcolsep}{3.6pt}
\begin{tabular}{@{}lccccccc|ccccccc|c@{}}
\toprule
Setting & Disc. & Lid. & Gath. & Pinch. & Fill.$^\dagger$ & Pack.$^\dagger$ & \textit{Avg} & Relay.$^\dagger$ & Sort.$^\dagger$ & Stor.$^\dagger$ & Bag. & Extr.$^\dagger$ & Pour.$^\dagger$ & \textit{Avg} & \textit{All}\\
\midrule
400 h & 10 & 7 & 27 & 0 & 13 & 23 & 13.3 & 40 & 53 & 39 & 9 & 54 & 16 & 35.2 & 24.2\\
2,000 h & 26 & 44 & 38 & 21 & 34 & 31 & 32.3 & 46 & 62 & 63 & 31 & 74 & 41 & 52.8 & 42.6\\
10,000 h (full) & 64 & 47 & 77 & 38 & 56 & 63 & 57.5 & 61 & 88 & 79 & 55 & 84 & 69 & 72.7 & 65.1\\
\midrule
\multicolumn{16}{@{}l}{\textit{Ablations of the 400-hour model}}\\
w/o demo wrist views & 11 & 13 & 11 & 0 & 9 & 17 & 10.2 & 37 & 42 & 21 & 12 & 23 & 13 & 24.7 & 17.4\\
w/o demo state & 7 & 8 & 23 & 3 & 6 & 11 & 9.7 & 36 & 45 & 28 & 7 & 46 & 12 & 29.0 & 19.3\\
w/o demonstration & 14 & 0 & 20 & 0 & 10 & 7 & 8.5 & 0 & 25 & 0 & 10 & 5 & 17 & 9.5 & 9.0\\
w/o pretraining & -- & -- & -- & -- & -- & -- & -- & 6 & 0 & 15 & 8 & 37 & 8 & 12.3 & --\\
\midrule
\multicolumn{16}{@{}l}{\textit{Ablations of the 10,000-hour model}}\\
w/o rollback & 51 & 33 & 75 & 27 & 51 & 42 & 46.5 & 60 & 85 & 77 & 35 & 46 & 67 & 61.7 & 54.1\\
w/o demonstration & 13 & 0 & 31 & 23 & 4 & 11 & 13.7 & 44 & 23 & 68 & 28 & 14 & 55 & 38.7 & 26.2\\
\bottomrule
\end{tabular}
\end{table*}

\begin{figure*}[t]
\centering
\includegraphics[width=\linewidth]{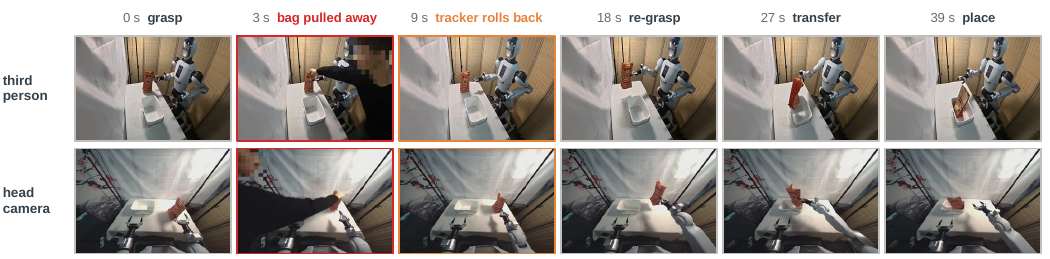}
\caption{\textbf{Recovery after a disturbance} (real robot, third-person and head views). At 3\,s the bag is pulled out of the gripper and moved; the tracker returns the window to the grasp phase, and the policy grasps again, transfers and places the bag.}
\label{fig:qual}
\end{figure*}

\begin{table}[t]
\centering
\caption{\textbf{Arm-joint error with the adapter off and on} ($10^{-3}$\,rad, 100 held-out chunks per source). $\Delta$: paired reduction, with 95\% bootstrap interval over query recordings (Rec.).}
\label{tab:adapter_mae}
\footnotesize
\setlength{\tabcolsep}{3pt}
\begin{tabular}{@{}lccccc@{}}
\toprule
Source & Rec. & Off & On & $\Delta$ [95\% CI] & Rel. \\
\midrule
UMI & 25 & 46.8 & 37.3 & 9.5 [5.1, 14.5] & 20\% \\
Human (ego2sim) & 37 & 50.2 & 42.1 & 8.1 [5.0, 11.6] & 16\% \\
Astribot S1 & 38 & 47.0 & 33.5 & 13.5 [10.1, 16.8] & 29\% \\
Spirit Moz1 & 48 & 48.1 & 29.8 & 18.3 [12.4, 24.5] & 38\% \\
\bottomrule
\end{tabular}
\end{table}

\begin{table}[t]
\centering
\caption{\textbf{SRA \vs full fine-tuning} on the same 200 pairs per source. Success (\%) with demonstrations from the source on seen tasks (tasks of the pairs, new scenes) and unseen tasks (Task Generalization).}
\label{tab:sft_vs_sra}
\footnotesize
\setlength{\tabcolsep}{3pt}
\begin{tabular}{@{}lcccc@{}}
\toprule
& \multicolumn{2}{c}{Seen tasks} & \multicolumn{2}{c}{Unseen tasks} \\
\cmidrule(lr){2-3}\cmidrule(lr){4-5}
Model & Human & UMI & Human & UMI \\
\midrule
Frozen (no adaptation) & 19.0 & 24.5 & 29.3 & 12.3 \\
Full fine-tuning       & 58.3 & \textbf{78.0} & \phantom{0}9.3 & 13.5 \\
SRA (ours)             & \textbf{62.8} & 75.5 & \textbf{49.3} & \textbf{46.3} \\
\bottomrule
\end{tabular}
\end{table}

\subsection{Source-Routed Adapters}
\label{app:exp_sra}

\para{Protocol of \cref{tab:sra}.} For each source and task, five source demonstrations, recorded or generated in scenes that differ from the test scenes, are used for the 100 trials, as in \cref{app:exp_benchmark}.

\para{Adapter off \vs on.} Beyond success, we measure what the adapter changes in the predicted actions. For each source we take 100 held-out query chunks (2\,s, 14 arm joints), give the policy the matching demonstration from that source, and predict once with the adapter off and once with it on; for the two robot sources the switch covers both the SRA and the embodiment adapter. \Cref{tab:adapter_mae} shows that the adapter lowers the arm-joint error by 16--38\% for every source, also on tasks outside its training pairs (UMI 41.3 to 36.0, human videos 66.4 to 53.2, $10^{-3}$\,rad). With the executor's own demonstrations the routing leaves the adapter inactive, and the output equals that of the frozen model, as the construction requires (\cref{app:sra}).

\para{SRA \vs full fine-tuning.} \Cref{sec:plugins} keeps $\theta$ frozen because fully fine-tuning it on a few hundred pairs risks overfitting. We test this for human videos and UMI by fine-tuning all 2.46B parameters of $\theta$ on the same 200 pairs as the SRA. Both are evaluated with demonstrations from the source on \emph{seen} tasks, four of the 20 tasks of the pairs in new scenes, and on \emph{unseen} tasks, the six Task Generalization tasks. \Cref{tab:sft_vs_sra} shows that full fine-tuning matches the SRA on seen tasks (58.3 \vs 62.8\% for human videos, 78.0 \vs 75.5\% for UMI), but on unseen tasks it drops to 9.3 and 13.5\%, close to or below the frozen model, whereas the SRA reaches 49.3 and 46.3\%. With all parameters free, the 200 pairs are fitted as behavior on their own tasks; the SRA, which changes only how the source's demonstration tokens are read, carries over to new tasks and leaves the executor's own demonstrations untouched.

\subsection{Progress Tracking and Recovery}
\label{app:exp_tracking}

\para{Window coverage.} For \cref{tab:align}, annotators mark, for every execution frame at 1\,Hz, the demonstration time that shows the same phase of the task, on 1,000 demonstration--execution pairs across sources and tasks. A frame is covered if this time lies inside the window $[\taudem_t-1\,\text{s},\ \taudem_t+3\,\text{s}]$ placed by the tracker run online, as deployed (\cref{app:tracking_forward}). Two annotators labeled 10\% of the pairs independently, with a median disagreement of 0.23\,s. The tracker uses the same hyperparameters for all sources and tasks.

\para{Recovery.} \Cref{fig:qual} shows a real-world rollout in which a person pulls the bag out of the gripper and moves it. The tracker rolls the window back to the grasp phase (\cref{app:rollback}), and the policy grasps the bag again and completes the task, without any recovery data. At 10K hours, disabling rollback costs 11.0 points (\cref{tab:ablation_pertask}). Further rollouts, rollback events and failure cases are shown on the \href{https://any2act.github.io}{project page}.

\section{Discussion}
\label{app:discussion}

\para{In-context learning and task data.} On Task Generalization, \ours uses no task data: each task is performed from a single demonstration with the pretrained weights. Skill Extension asks for skills beyond pretraining; there every method, including the baselines, receives executor data of the task, and \ours receives less than the VLAs (20 \vs 50 demonstrations per task). The demonstration remains essential after this adaptation: without it, a language-conditioned model fine-tuned on the same 20 demonstrations falls from 35.2\% to 9.5\% at 400 hours, and to 0\% on Relaying and Storing (\cref{tab:ablation_pertask}).

\para{Cost of a new source.} A translation pipeline is built once per source and reused for every task. Its adapter is trained once, on 200 pairs from 20 tasks disjoint from the evaluation tasks, for 500 steps with the pretrained model frozen (\cref{sec:exp_sra}). Because each adapter acts only on its own source's demonstration tokens, adding a source leaves the other sources and the executor's own demonstrations exactly unchanged (\cref{app:sra}), and unlike full fine-tuning it carries over to unseen tasks (\cref{app:exp_sra}).

\para{Robustness of progress tracking.} The tracker only needs to keep the correct phase of the demonstration inside the window; the policy, trained on perturbed and time-rescaled windows, locates its progress within it (\cref{app:tracking_tolerance}). The window tolerates a lag of up to 3\,s, and rollback returns it to an earlier phase after a failed step, so pauses, retries and speed differences between demonstrator and executor are handled without a learned aligner per source (\cref{tab:align}).

\para{Other executors.} The representation is defined by the executor, not by the Galbot G1: any robot with routinely collected teleoperation data provides its own robot-to-robot pairs. On RoboTwin 2.0, \ours is post-trained for RoboTwin's simulated dual-arm robot and outperforms Zero-WAM on six of the seven unseen tasks (\cref{tab:robotwin}); conversely, two other real robots serve as demonstrators of the Galbot G1 (\cref{tab:sra}).

\para{Limitations and future work.} Two directions remain open. First, the interface to an LLM agent is still narrow: the agent passes one demonstration of the intended task, whereas a more general interface would let it compose demonstrations of several subtasks, edit them, or generate them for tasks nobody has shown, as our simulation source begins to do. Second, the unified representation conveys what a demonstration looks like and how the arms move, but not how hard the hands press. Contact-rich tasks, such as inserting, wiping or tightening, depend on forces and touch that neither views nor joint trajectories record. Admitting force and tactile demonstrations, and recovering them from human videos through full 3D modeling of hand--object interaction rather than the object-level contacts ego2sim keeps today, would let a demonstration specify not only the motion but also the contact.

\para{Broader impact.} Following demonstrations from human videos makes it easier to instruct robots without programming, which can make assistive robotics more accessible. The same ease raises the question of whose demonstrations a robot should follow; deployment should restrict the accepted sources and keep a human able to stop the robot at any time. All human videos used in this work were recorded by consenting participants, and no faces are stored in the released material.